\documentclass[10pt,twocolumn,letterpaper]{article}

\usepackage[]{iccv}      
\usepackage{float}

\newcommand{\modelname}{P2P0.3}

\usepackage[T1]{fontenc}

\definecolor{iccvblue}{rgb}{0.21,0.49,0.74}
\usepackage[pagebackref,breaklinks,colorlinks,allcolors=iccvblue]{hyperref}

\def\paperID{*****} 
\def\confName{ICCV}
\def\confYear{2025}

\title{Learning to play: A Multimodal Agent for 3D Game-Play}

\author{Yuguang Yue
\and
Irakli Salia
\and
Samuel Hunt
\and
Christopher Green
\and
Wenzhe Shi
\and
Jonathan J Hunt \\
Player2\\
{\tt\small \url{https://player2.game}}
}

\begin{document}
\maketitle

\begin{abstract}
We argue that 3-D first-person video games are a challenging environment for real-time multi-modal reasoning. We first describe our dataset of human game-play, collected across a large variety of 3-D first-person games, which is both substantially larger and more diverse compared to prior publicly disclosed datasets, and contains text instructions.
We demonstrate that we can learn an inverse dynamics model from this dataset, which allows us to impute actions on a much larger dataset of publicly available videos of human game play that lack recorded actions.
We then train a text-conditioned agent for game playing using behavior cloning, with a custom architecture capable of realtime inference on a consumer GPU. We show the resulting model is capable of playing a variety of 3-D games and responding to text input. 
Finally, we outline some of the remaining challenges such as long-horizon tasks and quantitative evaluation across a large set of games. The game-playing videos from the model can be found at \url{https://blog.player2.game/p/pixels2play-v03-text-conditioned}.
\footnote{This workshop paper includes work previously published in \cite{yue2025pixels} but expands with significant new methodology and new and expanded results.}
\end{abstract}

\section{Introduction}

Artificial intelligence (AI) has been applied to gameplay since its inception \citep{turing1953digital}.  
Human performance in video games correlates with intelligence \citep{kokkinakis2017exploring,peters2021construction}. Games provide a cheap, safe, and quantifiable environment for evaluating new approaches \citep{costarelli2024gamebench, wang2023voyager}.

Recently, large language models (LLMs) such as ChatGPT \citep{brown2020language} have ushered general-purpose AI into daily life.  
Most commercial LLM offerings are now multi-modal visual language models (VLMs), accepting images as input. However, even with latency and cost constraints removed, current visual language models (VLMs) perform poorly at game control, for example none can finish the first level of the 1996 shooter Quake \citep{zhang2025videogamebenchvisionlanguagemodelscomplete}.
This shows that playing video games provides a challenging domain for multimodal real-time control \citep{hafner2023mastering}. Compared to other domains such as robotic manipulation or self-driving, for which many datasets are single-purpose \citep{delmerico2019we,yu2020bdd100k}, video games require modeling a much larger variation in behavior, objectives, and physics.

In this work, we focus on 3-D first-person view games. This provides some level of commonality across games to support generalization (for example spatial reasoning) while still providing a diverse set of tasks. Different games (and even game play within the same game) have vastly differing objectives, limits on control (e.g.\ first-person shooter vs. racing game) and a variety of physics and rendering engines.

\subsection{Contributions}

Our contributions in this work are as follows:

\begin{itemize}
    \item We describe the largest publicly disclosed dataset of high-fidelity recordings of human video game play. This dataset contains a wide variety of 3-D games. The dataset is also annotated with text descriptions of the behavior and the world that can be used for conditional behavior generation.
    \item We introduce a multi-game inverse dynamics model (IDM) trained on the above dataset that allows us to impute low-level actions on an even larger dataset of publicly available human gameplay (e.g.\ livestreamers playing games, etc.).
    \item We describe a novel architecture that allows us to train a text-conditioned model-free policy on the above data set that is capable of running in real time on a high-end consumer GPU.
    We describe and present empirical evidence for many of our model architecture choices.
    \item We demonstrate the model is capable of playing a variety of single games and responding to text conditioning.
    \item We outline the challenges that remain to provide consumer-applicable behavior models in general 3-D gameplay.
\end{itemize}

\section{Related work}

The literature on AI in game play is large; here we focus on recent work on video games.

One major area of focus has been the use of reinforcement learning (RL). This requires instrumenting the game to extract a reward, such as the score or win/loss of the game. For that reason, this approach is typically limited to playing a single game. Most works used a model-free approach; \citet{hafner2025mastering, hafner2025training} are notable exceptions.  Many approaches also substitute engineered state representations for raw pixels, adding additional game-specific engineering \citep{tesauro1991practical}.  With abundant computation, RL can achieve superhuman play \citep{mnih2013playing,vinyals2019grandmaster,berner2019dota}, although the resulting policies often differ markedly from the human style.

Behavior cloning (BC) re-frames control as supervised learning.  \citet{pearce2022counter, kanervisto2025world} applied behavior cloning in a single game, while \citet{raad2024scaling} trained a single model to play multiple games. \citet{tuyls2023scaling, pearce2024scaling} have investigated the scaling laws of behavior cloning. \citet{farhang2024humanlike} used offline RL to generate human-like behavior, but in an engineered state space rather than from pixels. \citet{kanervisto2025world} used a single model to model both the world and behavior (but modeled a single game). \citet{yue2025pixels} trained a vision-only model that directly predicts keyboard and mouse actions for a range of 3-D games, without conditioning on text input.

\citet{baker2022video} used behavior cloning to train a model to play Minecraft. Most of the training data was ``unlabeled'' from online video sources. In order to use these data, this work first trained an inverse dynamics model (IDM) from the labeled data and then used this to impute labels on the unlabeled data. Our approach to using IDM to benefit from unlabeled videos is similar; however, we attempt a diverse set of games rather than a single game.  Similar ideas have recently become popular in robotics, where learning from unlabeled videos is now common \citep{ren2025videoworld}.  Most robotics papers generate \emph{latent} actions: an IDM infers latent controls that are trained to encode information, via a forward dynamics model, useful for predicting the next frame. Policies are first trained on those latent actions and later mapped to real motor commands \cite{ye2024latent,bjorck2025gr00t,liang2025clam}.

The works above are purely behavior generation, but not multi-modal control. In particular, they lack any way to steer the model, with the objective typically being generating human or game-winnning behavior. \citet{lifshitz2023steve} provide one approach to add text conditioning to a behavior model. One of the few works that we are aware of that collected a multi-game dataset and trained a model to play multiple games from pixels is \citet{raad2024scaling}. They did not disclose the size of the dataset (or model). In this work, we target a model architecture that can perform inference in real-time on a high-end consumer GPU.

Vision-to-action models are an exciting area of multi-modal robotics research \citep{team2025gemini, black2410pi0, renz2025simlingo}, mapping from language and vision input to actions. Although many robotics models generate high frequency actions, they operate on visual input at relatively low frequency (typically less than 4Hz) \citep{zitkovich2023rt}, which would be too slow for competitive reaction time in real-time gaming and most recent foundation models for robotics require (multiple) server class GPUs to perform inference. Although robotics applications are extremely challenging, along some dimensions, general game play can be considered more varied with a wider variety of objectives. \citet{zheng2025flare} used implicit world modeling in a latent space to improve the generalization of a VLA.

\section{Methods}

\subsection{Dataset}

We set out to collect a large-scale dataset of high-quality human game play across multiple 3-D games. The data consists of trajectories of game play with the input seen by the human (the pixels on the screen $o_i$) and actions taken by the human (keyboard and mouse actions) $a_i$. A single episode consists of the observations $(o_1, ... o_n)$ and the corresponding actions the human took $(a_1, ..., a_n)$. As in \citet{baker2022video}, we record at a frequency of 20 Hz.

We worked hard to ensure that the dataset is of high quality. We recruited annotators with significant game-playing experience. In addition, all new annotators and game data were initially reviewed manually to check their quality. Once an annotator and game tuple were considered of consistent quality, we then relied on random checks and automated checks to ensure consistency. We instructed annotators to record only gameplay in the 3-D world (e.g. avoid menu selection prior to the gameplay beginning).

We created a number of automated checks, such as checking that no key was held down throughout the recording and that some keys/mouse movements were present in the trajectory. In addition, we used a simple version of our model-free policy (described later) to compute the likelihood of the behavior seen in the trajectory. Trajectories with unusually low likelihoods (often indicating an anomaly, such as the annotator left the recording running after the game finished or recorded the wrong screen) or that violate automated checks were surfaced for manual review.

A common challenge of behavior cloning is the distributional shift. The dataset consists of trajectories generated under human behavior, but the trajectories generated under a model policy may result in a difference in state distribution, e.g.\ visiting states unseen in the training set. We sought to mitigate this issue in two ways: collecting a large dataset to cover as many states as possible and collecting ``correction'' trajectories. When collecting correction trajectories, we allow the policy (described below) to control the game, but we allow the human annotator to take control at any time (for example, if the policy gets stuck or deviates from desired behavior). We collect the full trajectory (including while under model control) but record at each timestep if the actions are human-generated. When training the policy, we do not weight the loss on the model-chosen actions.

The resulting dataset has approximately 7000 hours of high-quality human game-play (and growing). It is our intention to make this dataset available on a non-commercial basis for academic research.

\subsubsection{Unlabeled data}

For most popular commercial games, there is a vast amount of publicly available video recordings of game play \citep{fan2022minedojo}. However, there are a number of challenges with using these data: it is of highly varying quality and can be interspersed or overlaid with non-game data, it is recorded at many different resolutions, frequencies, and aspect ratios, and most crucially, the key/mouse actions taken by the player are not available.

We curated a dataset of unlabeled (meaning without key or mouse movement) trajectories of human gameplay. We used public data sources and commercially available visual language models (VLMs) to curate the dataset. We implemented a two-step filtering process. First, an initial filter is applied based on metadata such as the video's title, description, topic, and thumbnail image (when available). This step involves querying a commercial VLM to assess the relevance of the videos to our specified query. Second, the full video content is processed by the VLM to segment and remove non-gameplay scenes. We used a low framerate video to balance cost and filter quality. We generated queries to obtain videos based on a large set of popular game titles.

\subsubsection{Text annotation}

There has been increasing interest in vision-to-action models in the robotics community \citep{kim2024openvla, black2024pi_0}.
Video games provide a challenging and interesting domain for research in this area; in many games, there are numerous different strategies and choices that a text-conditioned model could choose to play \citep{fan2022minedojo}. 

We therefore sought to add text descriptions of the behaviors exhibited in our dataset. The primary approach we used was to use commercial VLMs to retrospectively annotate the videos. That is, the instructions to our annotators were to play the games as competent players in a variety of different styles and to vary their choices during the gameplay. Although VLMs play games poorly, we found that commercial VLMs were capable of retrospectively annotating the behavior during gameplay to a reasonable level of quality. 

However, a significant limitation of these services is their tendency to automatically compress videos to a lower frame rate. This downsampling can be detrimental for video game annotation, as it creates temporal misalignments: an instruction tied to a specific moment (e.g., turn right at frame $t$) quickly becomes inaccurate at a subsequent frame $t+1$. To mitigate this, we iteratively refined our prompts to elicit instructions that were less sensitive to precise timing. For instance, we prompted the model to generate annotations focused on descriptive commands (e.g., go down the skull gate) rather than directional commands (e.g., go left). Furthermore, we adjusted the prompt to filter out high-frequency, repetitive events, such as "shooting" in a first-person shooter or "eating small animals" in a snake-style game. Our objective was to develop a prompt that was detailed enough to be useful yet general enough to apply across a large set of games with minimal manual curation, thereby ensuring the scalability of our approach.
The annotation output format is as follows
\begin{verbatim}
{
"narrative": "<string>",
instructions": [
 {
  "start": "<timestamp>",
  "end":   "<timestamp>",
  "instruction": "<string>"
 },
 {
  "start": "<timestamp>",
  "end":   "<timestamp>",
  "instruction": "<string>"
 },
 ...
]
}
\end{verbatim}

In addition to annotating behavior, we also generated text descriptions of the game state, which could be used for a text-conditioned world model. Using this retrospective annotation approach, we also annotated the unlabeled videos.

For a small percentage of trajectories, we requested that players include a text description of their goal or behavior (for example, by stating their intended behavior in advance in our annotation tool). An area of future work is to record communication between cooperating players in multiplayer games as additional behavior annotations and to use the text annotations to generate text-conditioned behavior models while maintaining the constraint of real-time inference on a consumer GPU.

\subsection{Inverse Dynamics Model}

Unlabeled gameplay videos greatly outnumber curated demonstrations, so we rely on an Inverse Dynamics Model (IDM) to turn those videos into additional training data.
Two IDM approaches appear in the literature: a \emph{real-action} model that predicts explicit key/mouse actions \cite{baker2022video}, and a \emph{latent-action} model that predicts abstract action codes later mapped to real actions \cite{schmidt2023learning,ye2024latent,nvidia2025gr00t}.
We adopt the real-action variant for simplicity; a direct comparison with latent-action IDMs is left for future study.

Formally, the IDM is a classifier over the action at time $t$ given the surrounding image sequence:
\[
\tilde{a}_{t}\;\sim\;p_{\text{IDM}}\bigl(a_{t}\mid o_{1},o_{2},\dots,o_{t},\dots,o_{T}\bigr).
\]
We used an identical architecture for the IDM model as that used for the policy (next section). The only difference is that we modify the masking because the IDM model does not need to be causal (that is, it can observe future frames when imputing an action). This has the benefit that the optimizations or improvements made to the policy model translated without further manual effort into improvements to the IDM model as well.

Training minimizes cross-entropy between the predicted distribution $\tilde{a}_{t}$ and the ground truth action $a_{t}$ on the labeled dataset. In order to ensure it is robust to many variations observed in the unlabeled dataset, we use extensive data augmentation such as cropping, perturbing the color space, small rotations, etc.\ of the training data. After training the IDM model, we use it to impute actions for the unlabeled dataset.

\subsection{Policy model}\label{policy_model}

Here we detail our architecture and training procedure for a general game-playing policy, which we call Pixels2Play (\modelname).

The primary constraint we place on our model is the need for the model to be capable of running in real-time (20 Hz) on a high-end consumer GPU (Nvidia RTX 5090). The reason for this constraint is that commercial applications require the model to run on end-user hardware.

\begin{figure*}[htbp]
    \centering
    \begin{subfigure}[t]{0.68\linewidth}
        \includegraphics[width=\linewidth]{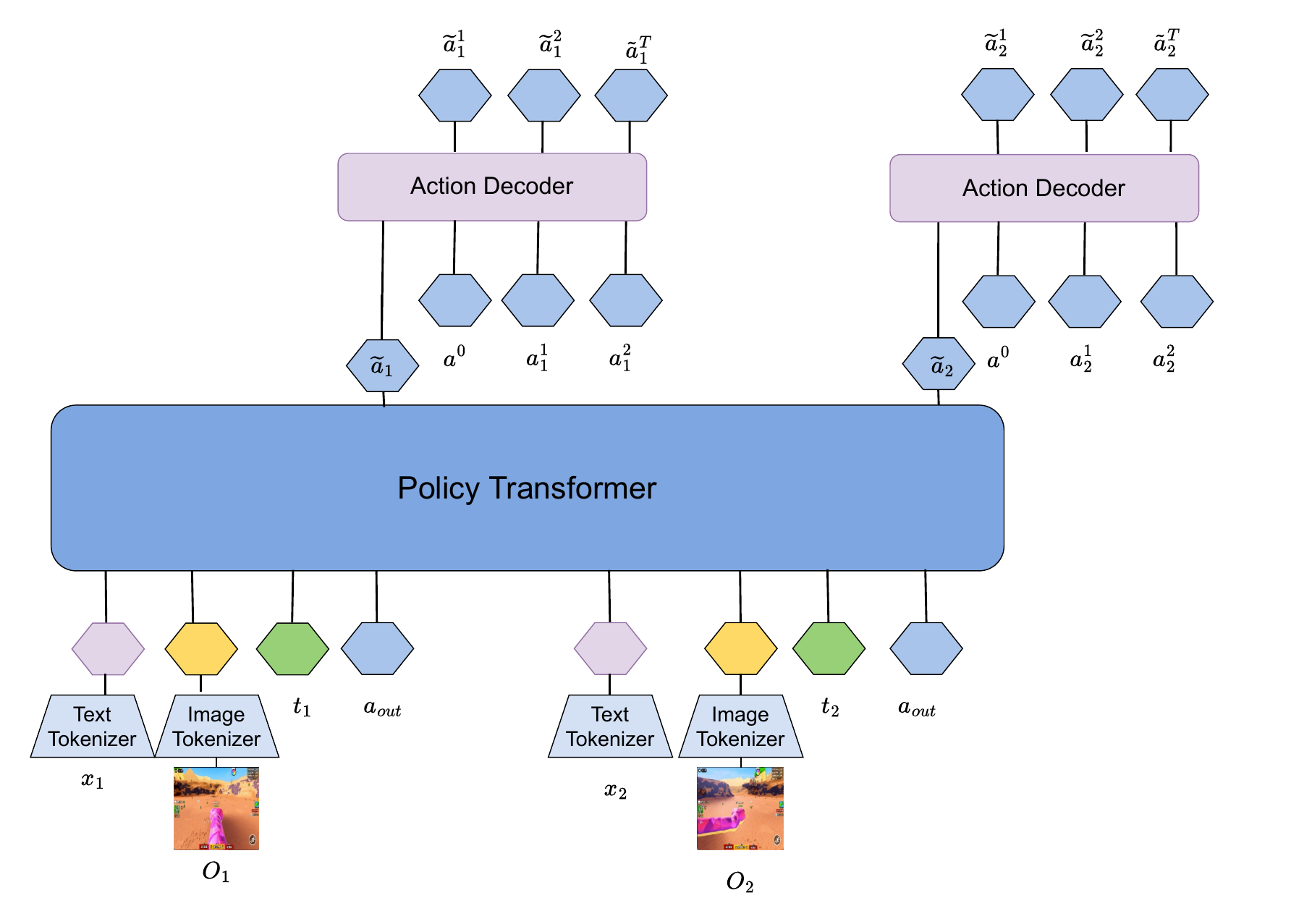}
        \caption{}
        \label{subfig:model}
    \end{subfigure}
    \hfill
    \begin{subfigure}[t]{0.25\linewidth}
        \includegraphics[width=\linewidth]{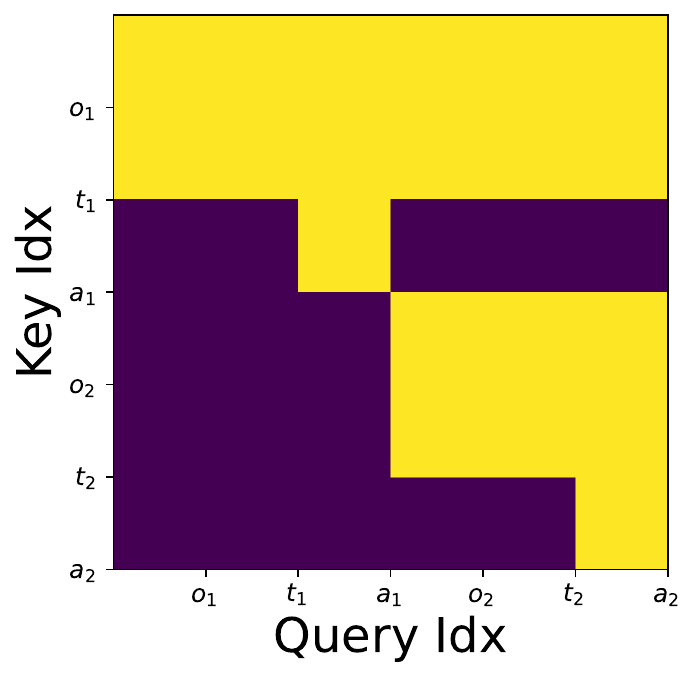}
        \caption{}
        \label{subfig:mask}
    \end{subfigure}
    \caption{
    \subref{subfig:model} Architecture of \modelname. The core policy transformer and action decoder are both decoder-only transformers.
Each timestep begins with a text token $x_i$. Since many frame do not contain a text annotation there is a learned default value $x_{null}$ inputed on these frames. This is followed by $\leq4$ image tokens from video frame $o_{i}$ 
followed by a learnable “reasoning’’ token $t_{i}$ that grants the model extra computation time. The policy transformer then outputs a single action token $\tilde{a}_i$. A smaller transformer, the action decoder, then auto-regressively transforms and samples the single action token into the full action space.
    \subref{subfig:mask}
    Attention mask used in our transformer policy (yellow denotes $1$ and blue $0$).
    Tick marks show the boundaries of successive inputs.  
The mask can be read by looking on the x-axis for the query and translating up to see the parts of the key masked out, e.g.\ tokens in $o_1$ can attend to $o_1$ and $t_1$ but not the future. Observe in step 2 that the prior action token $a_1$ is masked out.
}
    \label{fig:arch}
    \label{fig:mask}
\end{figure*}

In order to achieve this constraint, we do not use a pre-trained VLM as a starting point. Instead, we use a custom decoder-only transformer-based architecture (figure \ref{subfig:model}) to maximize inference efficiency. A typical VLM setup uses hundreds of tokens per image. While these allow for the model to be expressive and generalize, it significantly increases the number of tokens per timestep which decreases inference speed, drives VRAM usage, and limits the context length (history) that the model can retain. For that reason, we carefully design our architecture to minimize the number of tokens per timestep.

For the image encoding we use a pretrained image tokenizer based on the first 6 layers of a pre-trained efficientnet \citep{tan2019efficientnet} (as in \citet{pearce2022counter}) with a linear layer to project the encoding into a small number (1-4) tokens (we denote the number of image tokens $n_i$). We find that model performance is improved by not freezing the tokenizer but allowing the weights to update during training. However, we also find that starting from a pre-trained image tokenizer improves performance compared to initializing the tokenizer with random weights (figure \ref{fig:frozen}). As expected, we find that using more tokens per image improves performance (figure \ref{fig:n_img_tokens}).

Much prior work has focused on training a model to play a single game, which can often allow the use of a reduced action space \citep{baker2022video, pearce2022counter} that allowed these approaches to model actions as a single combinatorial categorical choice. However, as we wish the same model to master a variety of games, the action space is much larger (the entire keyboard and mouse space, and we allow up to 4 simultaneous keypresses and 2 simultaneous mouse actions). Modeling this with the same combinatorial approach would be infeasible. Therefore, we model the action distribution auto-regressively. To avoid increasing the number of tokens per timestep of the primary transformer, we use a smaller \textit{action decoder} that takes a single action token and decodes it auto-regressively into the full action space (somewhat analogous to the use of flow-matching or other approaches to generating a complex action space seen in robotics foundation models).

The policy transformer therefore only requires a small number of tokens per timestep: image tokens, one text conditioning token and one token for the action prediction. However, we found that model performance improves (at the cost of additional tokens) by introducing a ``reasoning'' token $t_i$ to allow the model an additional timestep for reasoning before the output of an action. The final result is $n_i + 2$ tokens per timestep.
We add a learned embedding to the input tokens indicating the type of token (image, text, reasoning, action out). In addition, at each layer of the transformer we add rotatory position embeddings \citep{su2024roformer}. Of course, during inference, we use key-value caching to reduce the computation complexity during inference. We use a sliding-window attention to maintain a maximum cache size and prevent the VRAM usage from growing indefinitely.

The policy model (unlike the IDM) must be causal (that is, it cannot make use of future observations to determine action). However, due to the design of the model, we do not use standard causal masking (Figure \ref{subfig:mask}). Since the tokenizer and reasoning steps will be computed at inference time in a single forward pass, we unmask reasoning step $t_i$ to $o_i$.

Behavior-cloning agents often suffer causal confusion \citep{de2019causal}: e.g.\ keys are often held for multiple frames, and the network may learn to copy the previous action rather than attend to pixels, this situation can be worse if the frequency of the video is high. We find that allowing model predictions at time $i$ to observe past action tokens $a_{<i}$ results in models that had strong offline metrics, but performed very poorly when evaluated online. 

To quantify this observation, we empirically estimated the causality of the model by permuting $50\%$ of observations in the validation set between different trajectories in a batch while keeping the action sequence unchanged. We then calculated the Kullback-Leibler divergence between the action predictions of the original sequence and the permuted sequence. While an imperfect estimator, a larger divergence indicates the model is making action choices based on the content of the observations rather than non-causally by using the prior action.

We find (Figure \ref{fig:prior-actions}) that the model performs better when masking out past action choices. Masking prior actions sacrifices optimality in certain cases.
For instance, in the FPS games where mouse movement is required, attending to prior actions is important for the model to adjust to different varying mouse sensitivity settings; a human player will calibrate their mouse movement based on past actions; without seeing earlier actions, the model cannot make such adjustments. 
Empirically, the policy remains adequate. Determining how to allow the model to observe past actions while maintaining causality is an area of active consideration.

We initially found that the models performed poorly due to a significant distributional shift between model training data and inference, which we traced to two factors: 1. During inference, no video compression takes place, but training data (for practical reasons) must be compressed. 2. The image resizing function differed between training (Python) and inference (Rust). The offline metrics looked promising, but when evaluated online, the models often failed to take any actions. We alleviated the distributional shift by using data augmentation to improve robustness, randomizing compression quality during video compression, and using the same resizing function between training and inference.

We have experimented with the use of sink tokens for sliding window attention \citep{barbero2025llms, xiao2023efficient}. In our approach, we learn a small number of key and value tokens at each self-attention layer, which are always inserted at the beginning of the key/value sequence when computing self-attention (including at inference time when using a key-value cache). We found that the sink token appears to have little effect (Appendix \ref{appendix:sinktoken}).

The majority of our experiments used an 11-layer, 2048 dimensional policy transformer. With image tokenizer and action decoder the total parameter count was approximately 400 million. We used mixed precision training \citep{micikevicius2017mixed} (weights represented as float32, activations as bfloat16, except in RMSNorm layers which used float32 precision). All training was performed on Nvidia 8xH100 and all evaluation used an Nvidia RTX 5090.

\subsubsection{Text annotation strategy}
\label{sec:annotation}

At each frame the model has an optional text annotation token (which can instead be set to $x_{null}$ to indicate no new text). We experimented with two types of text annotation strategies: the first strategy was only adding the text annotation once on the start timestamp, denoted the single-annotation-frame approach; and the second strategy was to repeat the text annotation for all the frames between the start frame index and the end frame index denoted repeat-annotation-frame approach. The advantage of the single-annotation-frame approach is to have better alignment between model training and inference - we can just add the text to the frame whenever the user gives instruction during inference without needing an explicit end time. In the repeat-annotation-frame approach, the user or system will need to determine explicitly when to end the text conditioning. On the other hand, the repeat-annotation-frame approach has a denser ratio of text annotations which can improve the text model with a limited amount of data. In our dataset, with the single-annotation-frame approach, only have $0.1\%$ of frames with text annotations whereas the repeat-annotation-frame approach has $\approx 10\% - 30\%$ frames with text annotations. 

As outlined in section \ref{sec:text-tokenizer}, we evaluated the model using 3 different text tokenizers.

\subsubsection{Evaluation}

During training, we can easily observe both training and validation loss. However, offline metrics may not correspond to the online performance of the model. A significant challenge for evaluation is that we wish our model to play a large variety of games. Instrumenting even a single game for automated performance evaluation is time-consuming. For this reason, currently, we are primarily limited to a qualitative evaluation of the model's performance.

We have developed two simple internal games. A simple racing game and a first-person shooter that we have instrumented to allow running synchronously with the model and provide quantitative evaluation. However, we found that these evaluations were still noisy, as small changes in the model or dataset composition can result in significant changes in a single game.

Automating model evaluation in a variety of environments is an area of active work. For now, we primarily rely on qualitative analysis of video recordings across a variety of games.

Evaluation of text conditioning is even more challenging. For some games, it is possible to reset the state to a specific initial condition. This allows sampling the generated behavior from the same state under differing text instructions. We use this approach to test the instruction-following capabilities of the model.

\section{Experiments}

We present a number of experiments. Due to space limitations, some results are left for the appendices. Appendix \ref{appendix:reasoning} shows the importance of reasoning tokens on model performance, Appendix \ref{appendix:sinktoken} shows that sink tokens do not have a significant impact on model performance and appendix \ref{appendix:causality} shows that the policy is non-causal when we do not mask out past actions.

\subsection{IDM model learns a lower perplexity than a causal policy}

We use the same model architecture for IDM and learning a causal policy. The only difference is that the IDM can observe future frames (they are unmasked) to determine what action took place and the IDM is not text-conditioned. As expected, Figure \ref{fig:idm} shows the IDM has a lower perplexity action loss, both in training and validation. This demonstrates that the IDM is making use of future frames to improve its prediction of the actions.

\subsection{Training with unlabelled data with IDM}

We show we can benefit from training using unlabelled game play videos (videos where the actions are not recorded), commonly available publicly. We have three stages of training to train with unlabelled data: We need to first train an IDM model on labelled data (figure \ref{fig:idm}); Then we train a policy model using a mixture of labelled data and unlabelled data with imputed labels generated from the IDM for the unlabelled data (analogous to LLM literature, we refer to this stage as ``pre-training''; Finally, we finetune the pretrained policy model on only the labelled data with a smaller learning rate. In these experiments, we had approximately 4x more unlabelled data (16000 hours), although we intend to increase this amount in the future.

\begin{figure}[htbp] 
    \centering 
    
    \begin{subfigure}{0.3\textwidth}
        \centering
        \includegraphics[width=\linewidth]{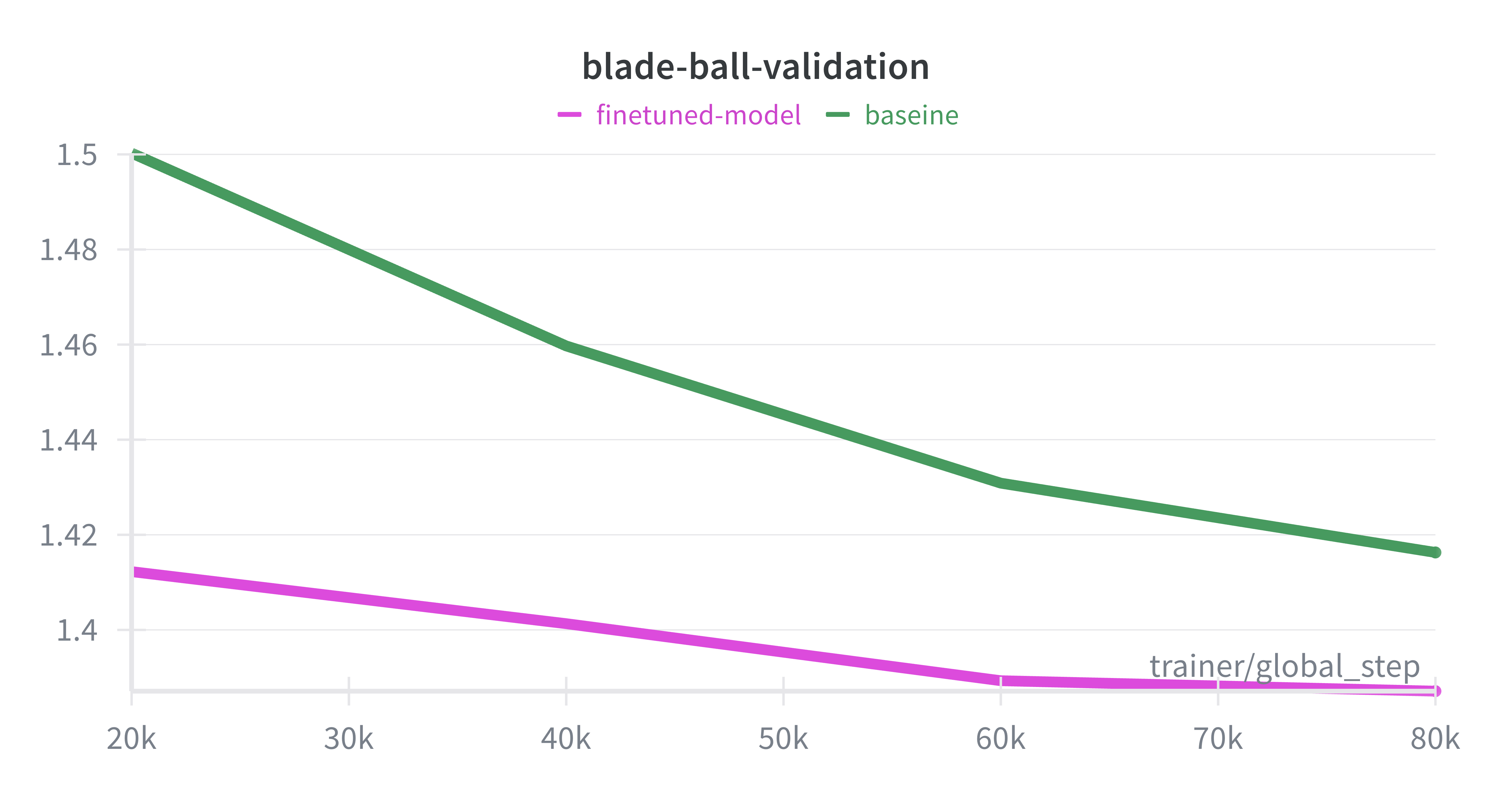}
    \end{subfigure}
    \begin{subfigure}{0.3\textwidth}
        \centering
        \includegraphics[width=\linewidth]{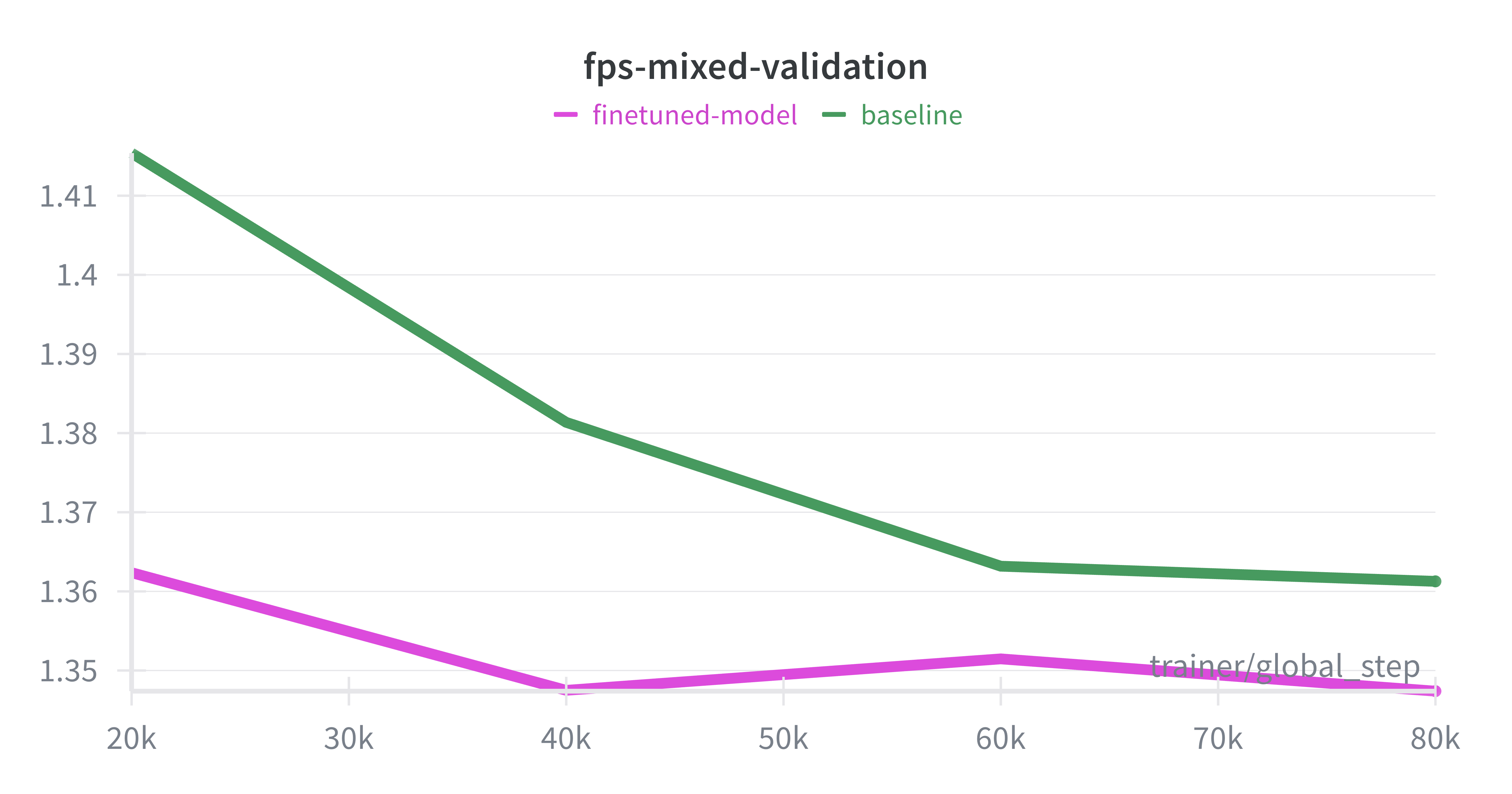}
    \end{subfigure}
    \begin{subfigure}{0.3\textwidth}
        \centering
        \includegraphics[width=\linewidth]{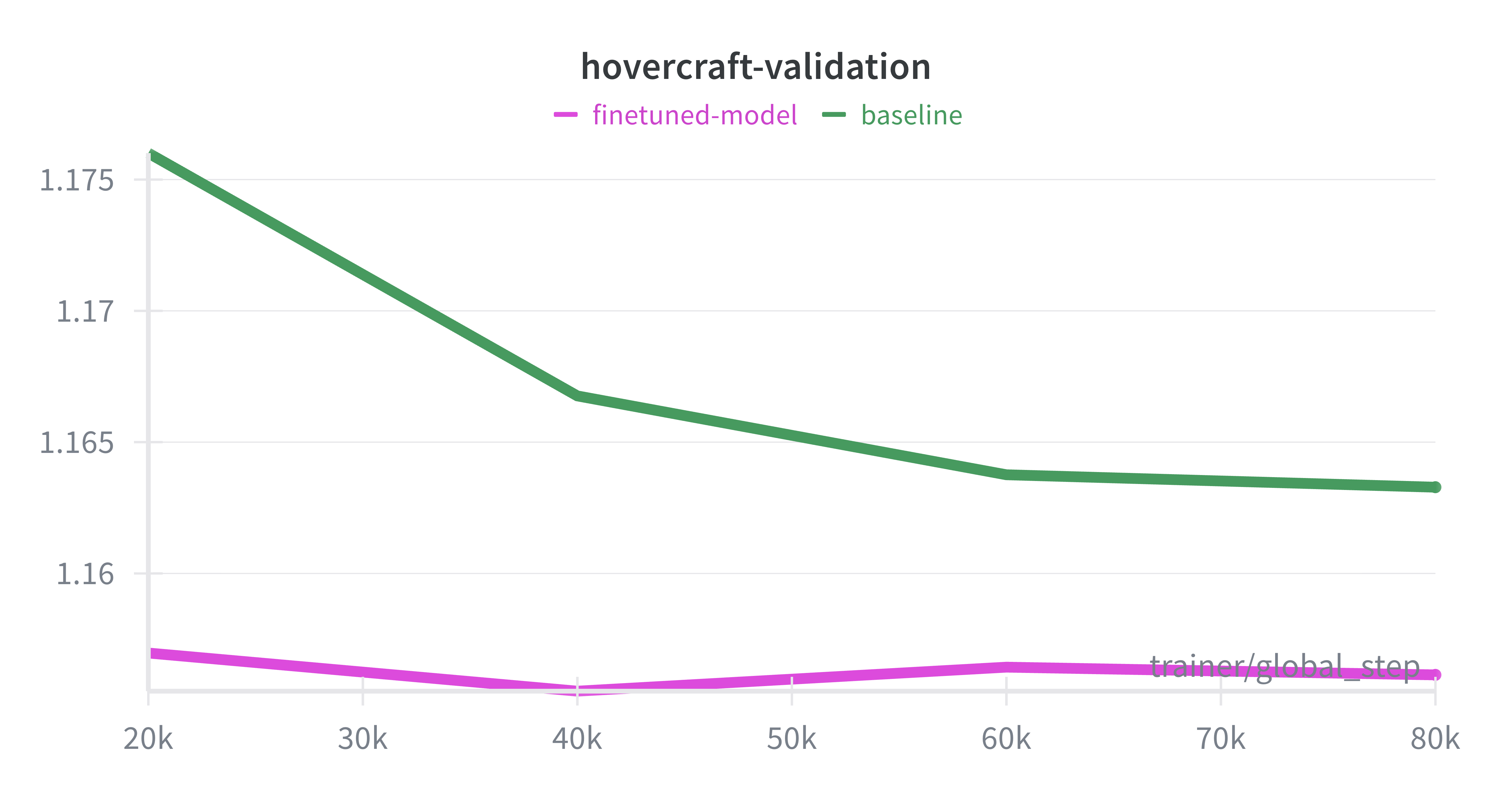}
    \end{subfigure}
    \caption{Validation metrics for finetuned models compared with baseline model trained only on labeled data. We consistently observe fine-tuning the pre-trained model results in improved performance, indicating the additional unlabelled data improves model generalization.}
    \label{fig:pretrain-finetune}
\end{figure}

After the pre-training stage the model has a significantly higher perplexity than one trained only on the high-quality labelled dataset (figure \ref{subfig:pretrain} and figure \ref{fig:pretrain-valid}). However, when the pre-trained model is fine-tuned using only labelled data it reaches significantly lower training (figure \ref{subfig:fine-tune}) and validation perplexity (figure \ref{fig:pretrain-finetune}), demonstrating the pre-training phase improves generalization.

\subsection{General game playing}

Currently, we have focused on simple Roblox games, older MS-DOS titles, and modern first person shooter and racing games. The Roblox platform has the advantage that, as it reduces the barriers to game design, it has a very wide variety of games. We are also learning MS-DOS games as part of a goal to use automatic evaluation in the future. In all games, we capture training data and evaluate directly on end-user computers with no instrumentation or modification to the games, with the games running in realtime.

\begin{figure}
    \centering
    \includegraphics[width=\linewidth]{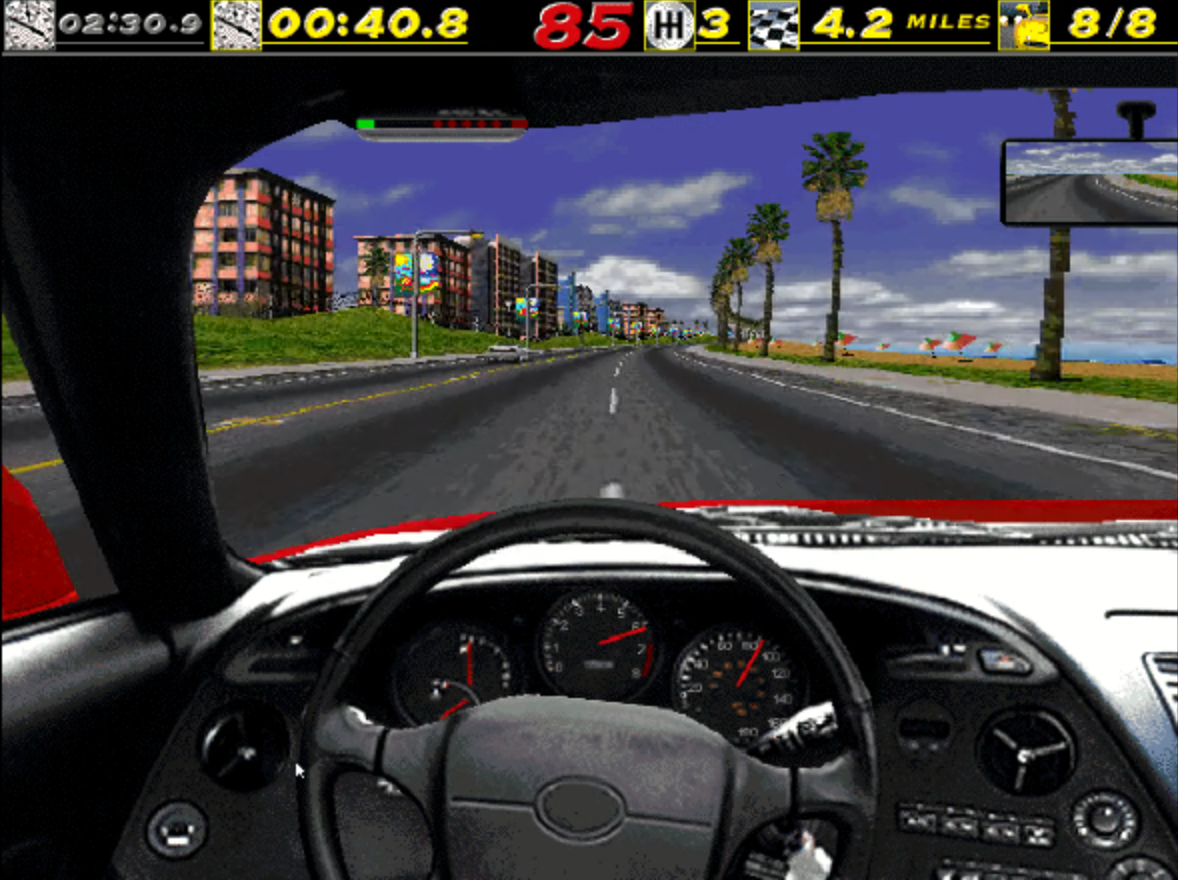}
    \includegraphics[width=\linewidth]{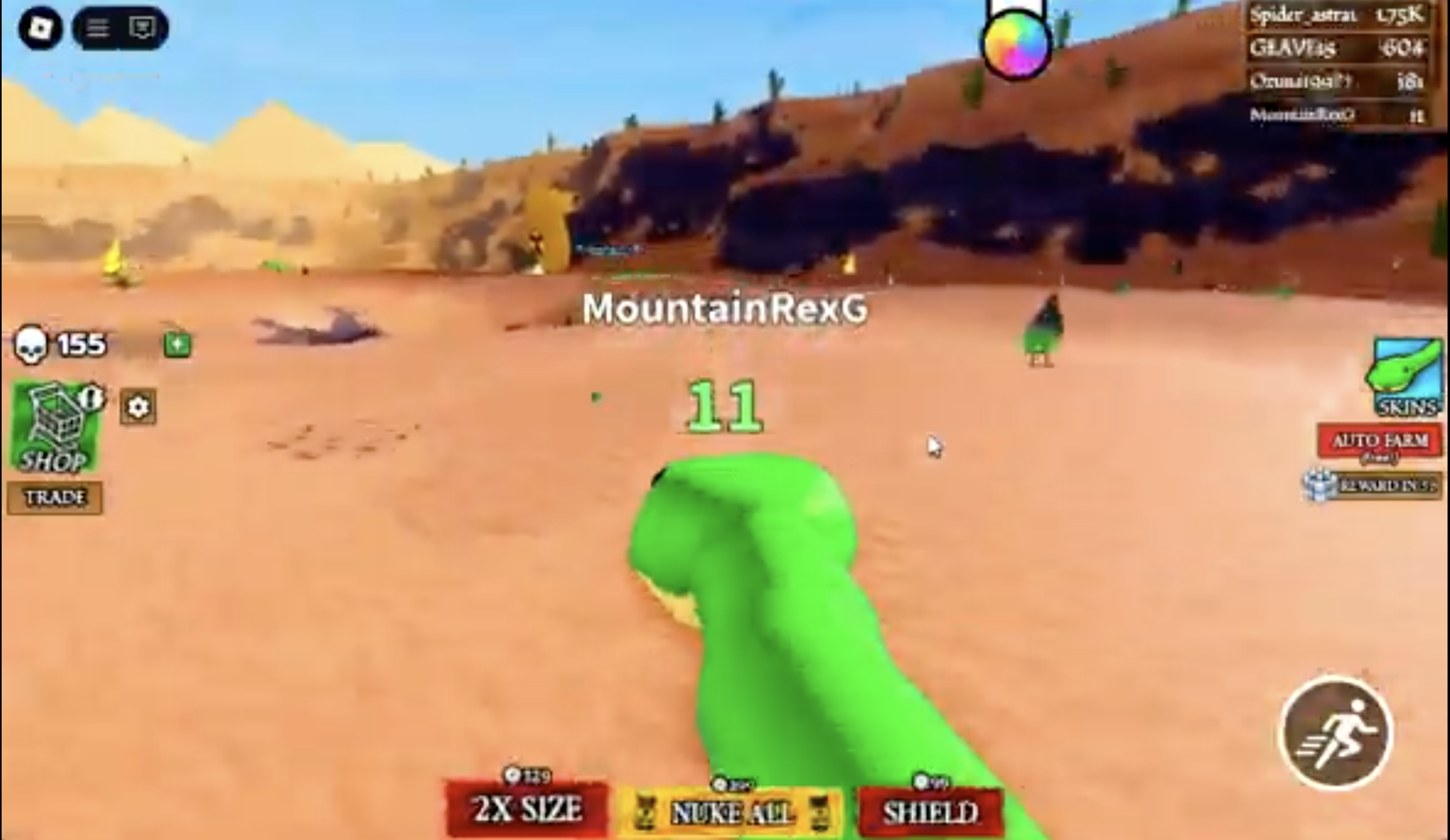}
    \includegraphics[width=\linewidth]{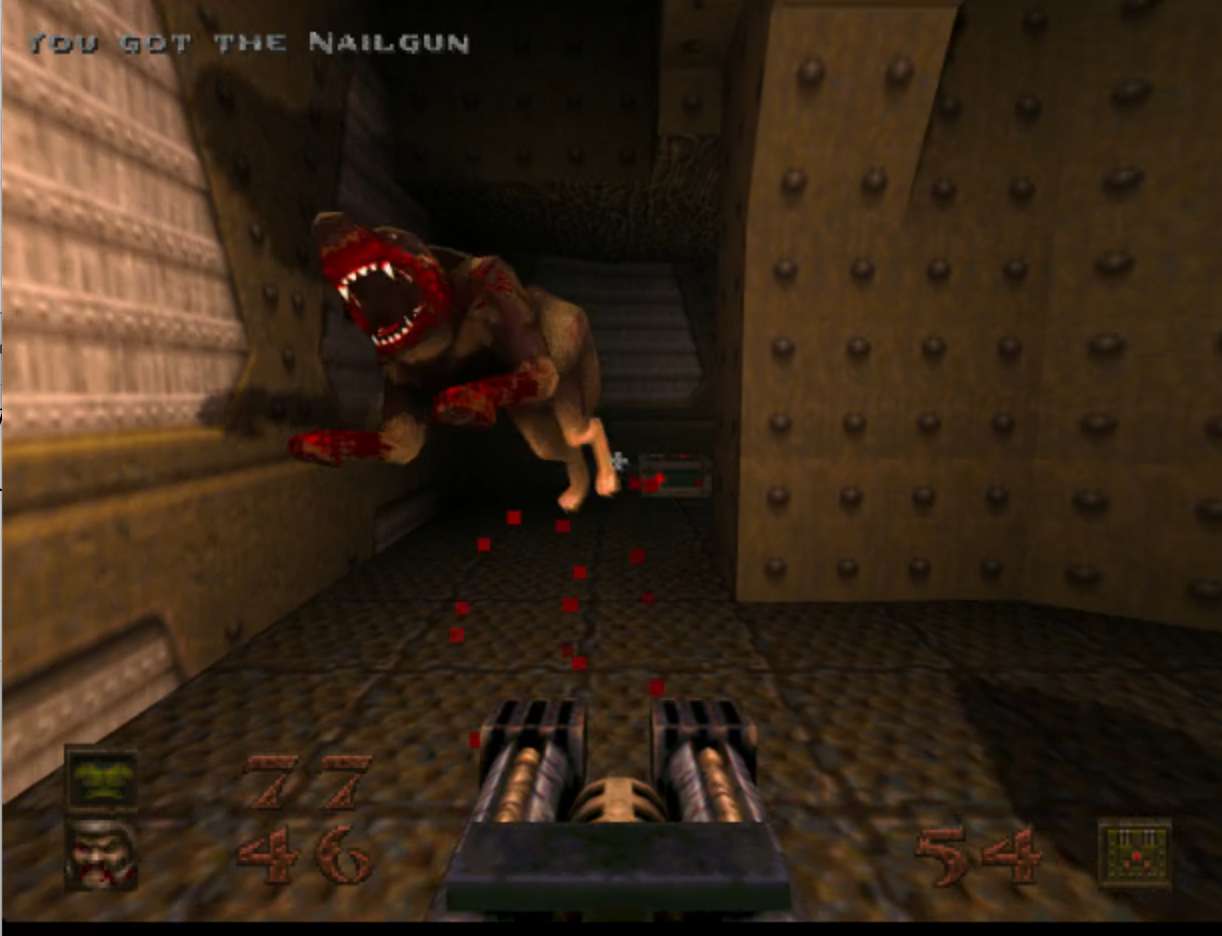}
    \caption{
    Examples of a MS-DOS game, Roblox,  and first person shooters \modelname~is currently capable of playing. For simple games such as MS-DOS Need For Speed or Roblox ``Be a Snake'' or simple FPS the model is capable of gameplay at the level of non-expert human. However, for games which require longer-term planning such the model struggles to complete levels. Video of gameplay can be seen view at \url{https://blog.player2.game/p/pixels2play-v03-text-conditioned}.
    }
    \label{fig:game}
\end{figure}

As discussed above, instrumented evaluation is an area of active work. Qualitatively, we find that \modelname\ is currently capable of playing simple games at the level of a novice human (Figure \ref{fig:game}), but performs poorly in more complex tasks or games that require longer-term planning.

\section{Text Conditioned Model Experiments}

\begin{figure}[t]
    \centering
    \begin{subfigure}{0.3\textwidth}
        \centering
        \includegraphics[width=\linewidth]{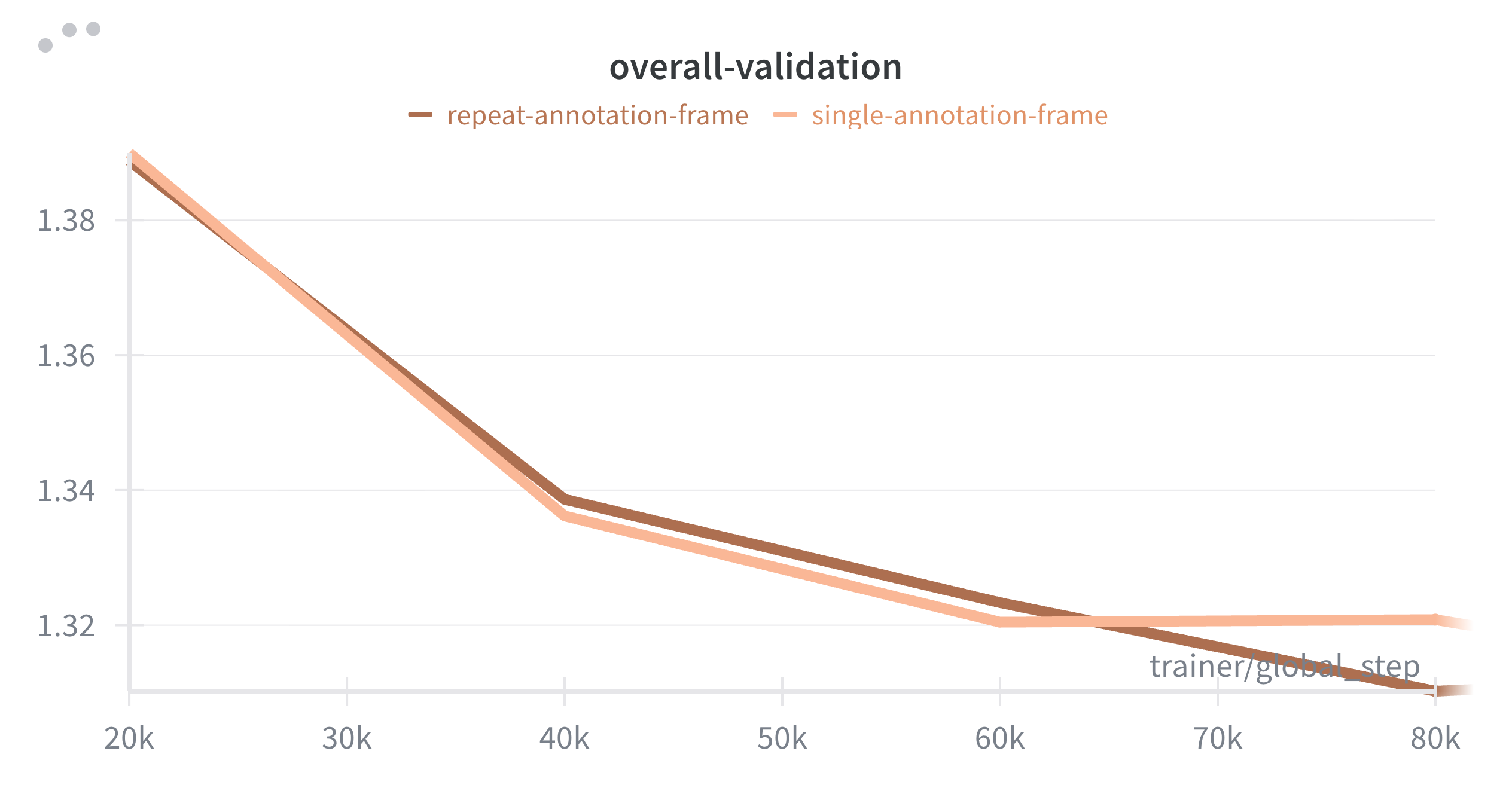}
        \caption{Validation perplexity for the two different text annotation strategies (see section \ref{sec:annotation})}
        \label{fig:annotatino-strategy}
    \end{subfigure}
    \begin{subfigure}{0.3\textwidth}
        \centering
        \includegraphics[width=\linewidth]{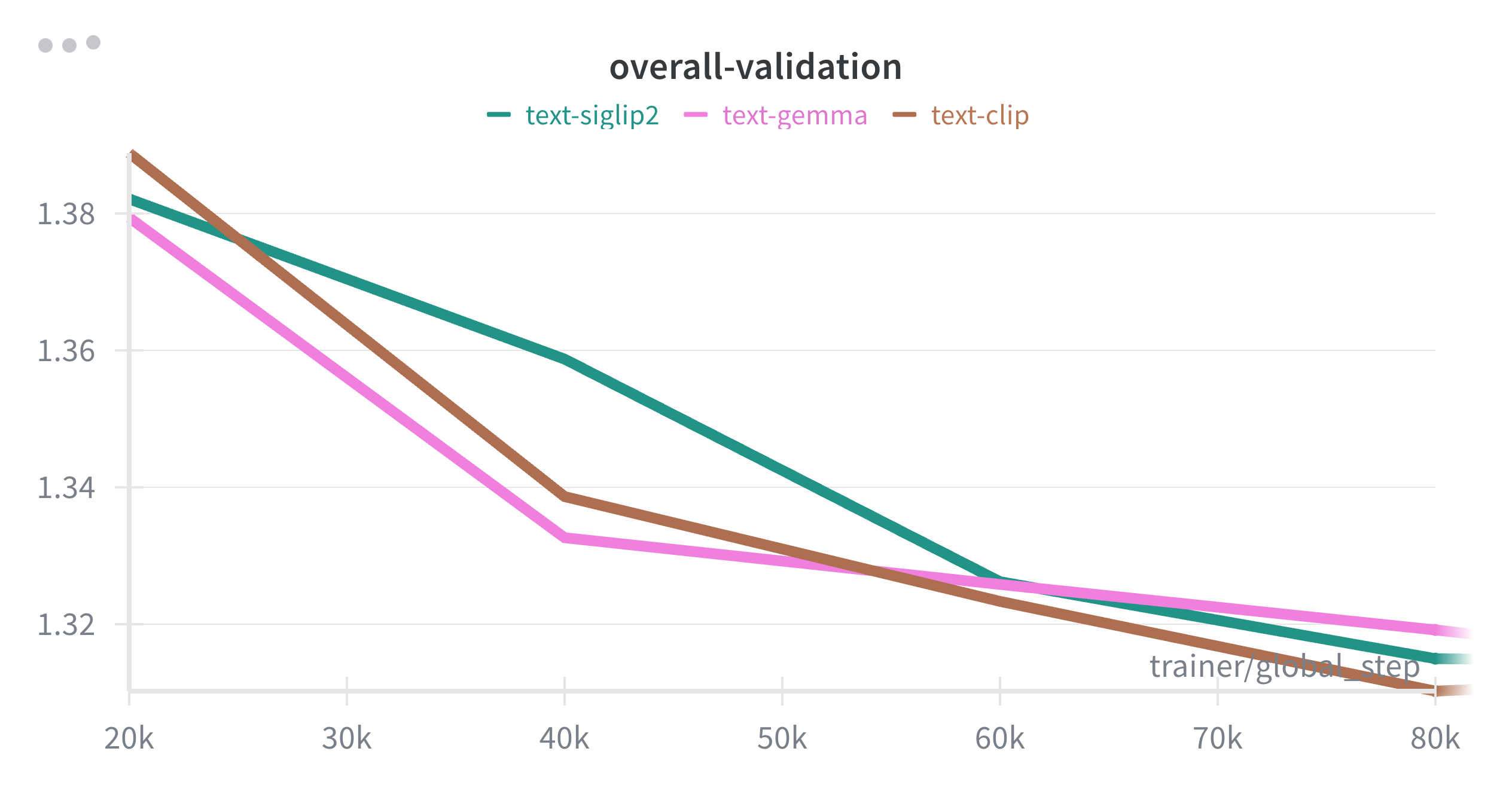}
        \caption{Validation perplexity using different text tokenizers. }
        \label{fig:tokenizer}
    \end{subfigure}
    \caption{Validation perplexity}
\end{figure}

From figure~\ref{fig:annotatino-strategy} we can see the repeat-annotation-frame approach has a slightly better validation loss, which is expected as it used more text annotation signals.
We repeat the text annotations a maximum number of times, then when issuing text conditioning during inference we always persist it for this maximum, rather than trying to determine a variable end time.

\subsection{Text tokenizer choice} \label{sec:text-tokenizer}
We experiment with three different text tokenizers: CLIP text tokenizer \cite{radford2021learning}, Gemma text tokenizer \cite{team2024gemma} and Siglip2 text tokenizer \cite{tschannen2025siglip2multilingualvisionlanguage}. We freeze the tokenizers during training and we obtain the instruction sentence embedding by averaging over the token embeddings. As shown in figure~\ref{fig:tokenizer}, the CLIP text tokenizer performed slightly better than the rest. 


\subsection{Qualitative evaluation of the text instructions}
To evaluate the impact of text instructions, we resume the game from a set checkpoint and observe the model behavior conditioned on varying text annotations qualitatively. We choose to use Quake and Doom for our qualitative evaluation because it is simple to set checkpoints in those single-player games, and it is easy to observe different model behaviors. 

\begin{figure}
    \centering
    \begin{subfigure}{0.43\textwidth}
        \centering
        \includegraphics[width=\linewidth]{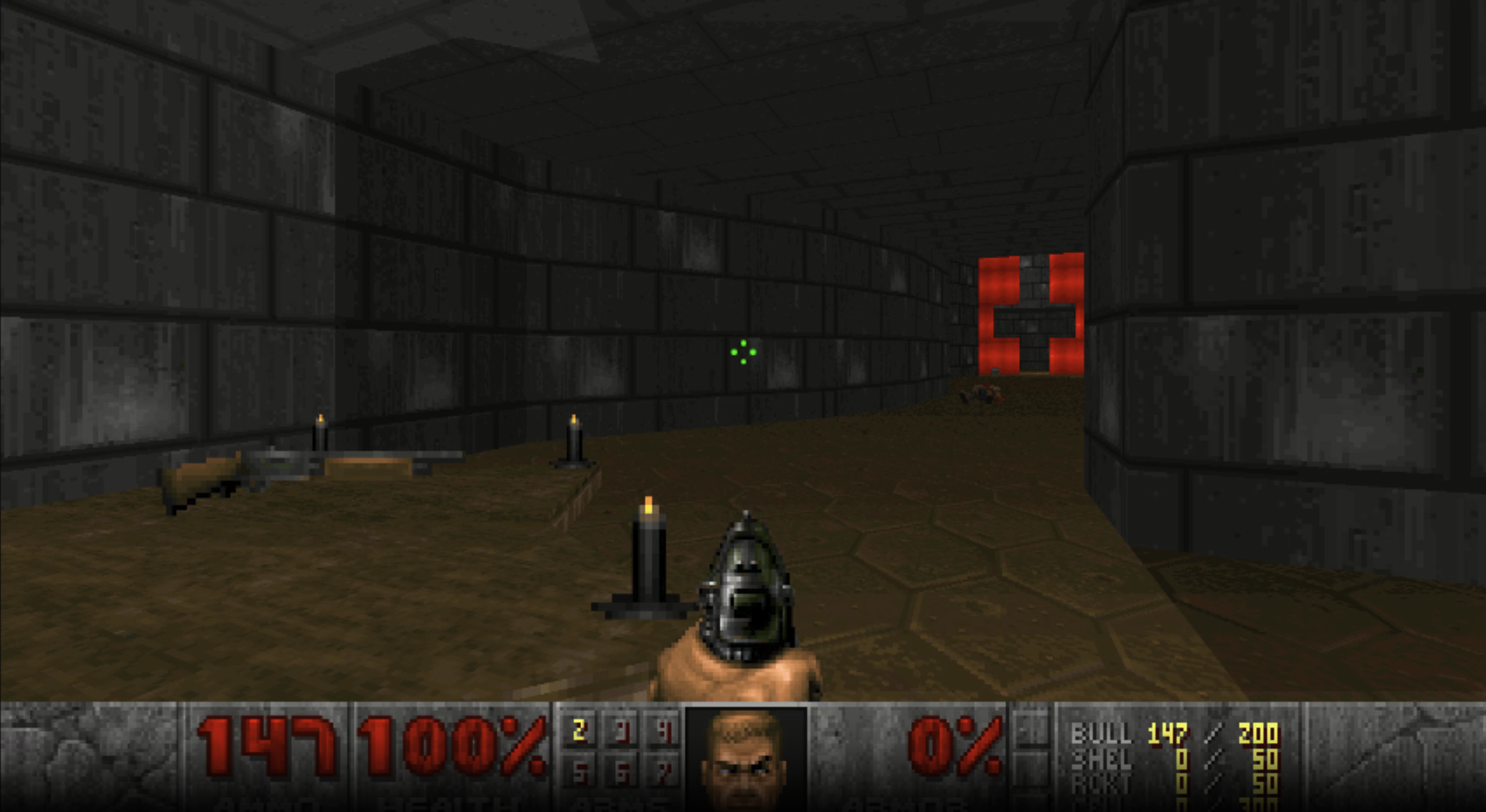}
        \caption{checkpoint screenshot of Doom}
        \label{fig:doom}
    \end{subfigure}
    \begin{subfigure}{0.43\textwidth}
        \centering
        \includegraphics[width=\linewidth]{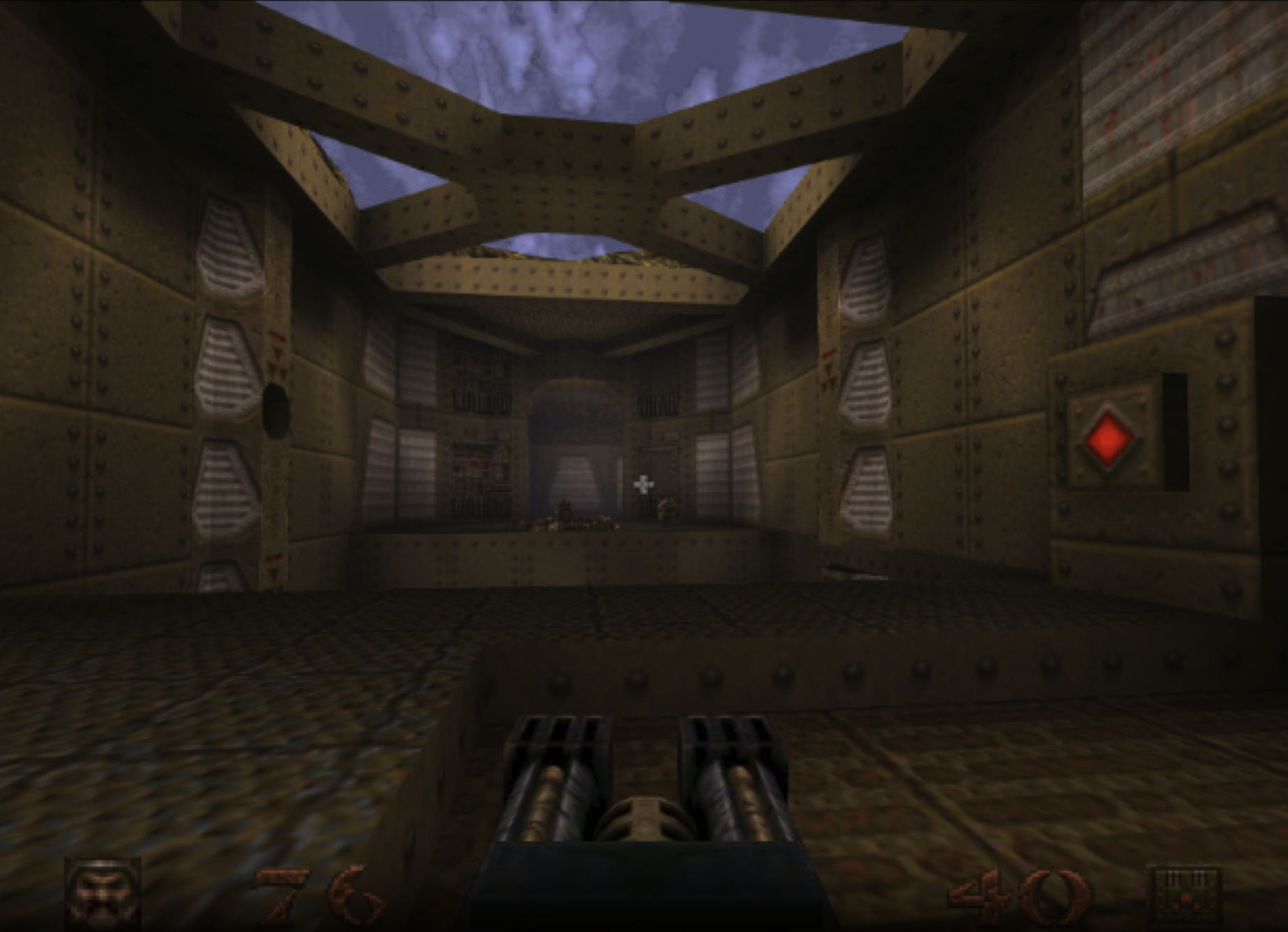}
        \caption{checkpoint screenshot of Quake}
        \label{fig:quake}
    \end{subfigure}
    \caption{To evaluate the text conditioning we start the model from the same checkpoint with differing text prompts.}
    \label{fig:checkpoints}
\end{figure}

We set the checkpoints at Quake and Doom as shown in figure~\ref{fig:checkpoints}. For the Doom checkpoint (figure~\ref{fig:doom}), there is a shotgun on the left and a red-cross door on the right. We tried three text instructions: (1) no text input; (2) ``pick up the shotgun''; (3) ``proceed to the red-cross door.'' For the Quake checkpoint (figure~\ref{fig:quake}), there is a red button on the wall and the player needs to go to the wall to press the button so that the bridge deploys. We tried two text instruction: (1) no text input; (2) ``move to the wall and press the red button".

We evaluated the model performance by counting how many times the model achieves the instructions in a short time duration (5-10 seconds). We performed 5 trials for each condition. The results are shown in table~\ref{table:eval}. These provide clear evidence that the model can follow the text instructions. 

\begin{table}[]
\begin{tabular}{|l|l|l|}
\hline
Doom                                                                                   & picked up shotgun  & reached door \\ \hline
no text input                                                                          & 1/5                & 1/5          \\ \hline
pick up shotgun                                                                        & 5/5                & 0/5          \\ \hline
\begin{tabular}[c]{@{}l@{}}proceed to the\\ red-cross door\end{tabular}                & 1/5                & 2/5          \\ \hline
Quake                                                                                  & pressed the button &              \\ \hline
no text input                                                                          & 0/5                &              \\ \hline
\begin{tabular}[c]{@{}l@{}}move to the wall\\ and press the red \\ button\end{tabular} & 2/5                &              \\ \hline
\end{tabular}
\caption{Success rate (out of 5 attempts) for each task evaluating text conditioning. We see clear evidence the model behavior is steered by the text conditioning.}
\label{table:eval}
\end{table}

\section{Discussion}

This paper outlines the challenges of text conditioned control in the diverse world of 3-D video games. We have outlined an approach to training a model-free policy using large-scale behavior cloning that is both text conditioned and runs in realtime on a consumer PC. More generally, we argue that the dataset and domain of 3-D games remains a challenging area for future work.

Currently, \modelname~handles a range of relatively simple 3-D titles with some level of text control.  The ongoing work focuses on two main fronts.  First, we continue to iterate on architecture and scaling, enlarging both the labeled and unlabeled corpora and increasing model capacity.  Second, we are extending the temporal window so that the agent can reason over much longer histories, a prerequisite for competent play in more complex games.



{
    \small
    \bibliographystyle{ieeenat_fullname}
    \bibliography{refs}
}

\appendix

\renewcommand{\thefigure}{A\arabic{figure}}

\onecolumn
\section{Additional figures}

\subsection{Frozen Image Tokenizer Performs Worse}

\begin{figure}[H]
    \centering
    \begin{subfigure}{0.48\textwidth}
        \includegraphics[width=\linewidth]{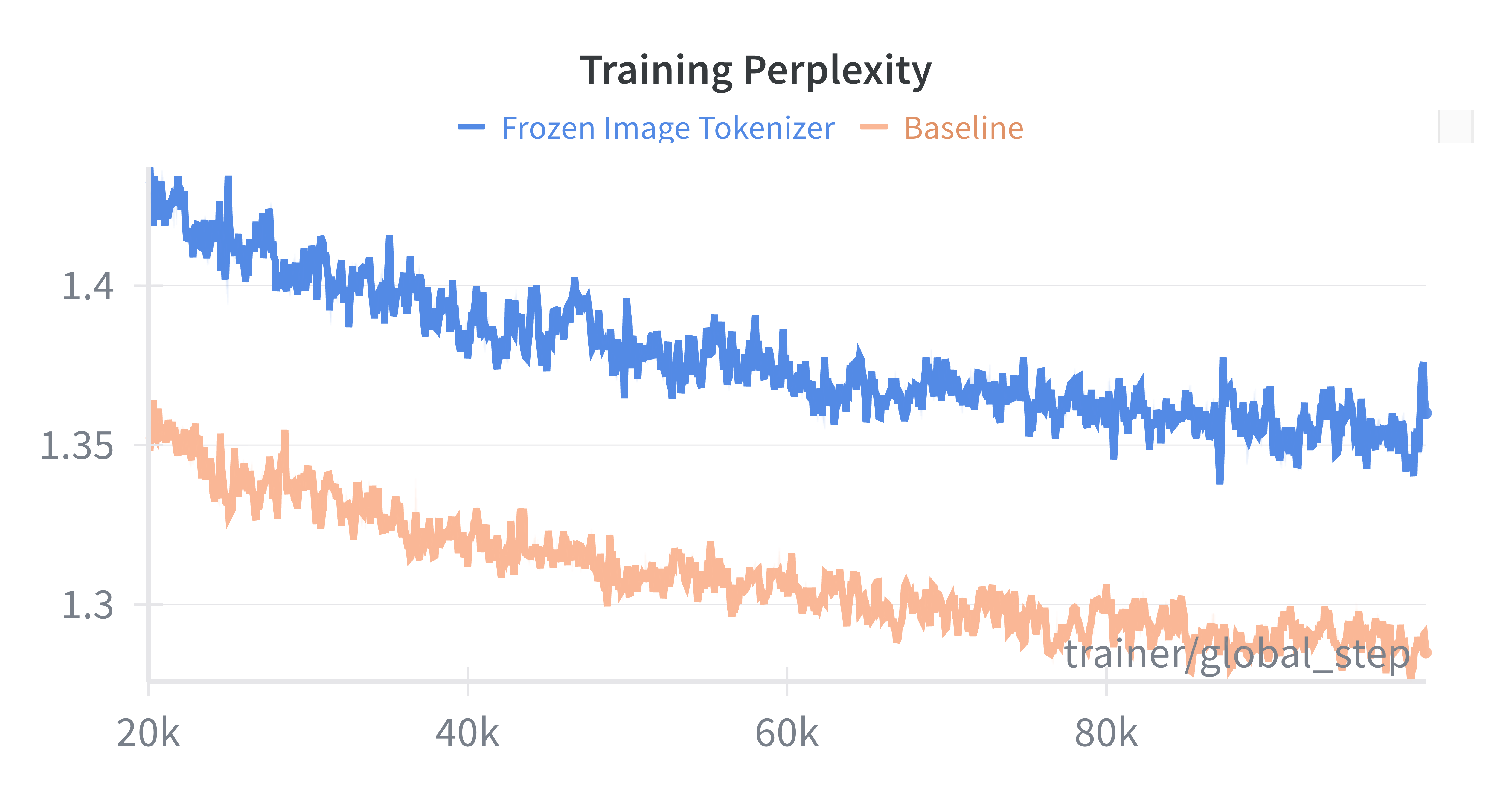}
        \caption{}
        \label{subfig:frozen_train}
    \end{subfigure}
    \begin{subfigure}{0.48\textwidth}
        \includegraphics[width=\linewidth]{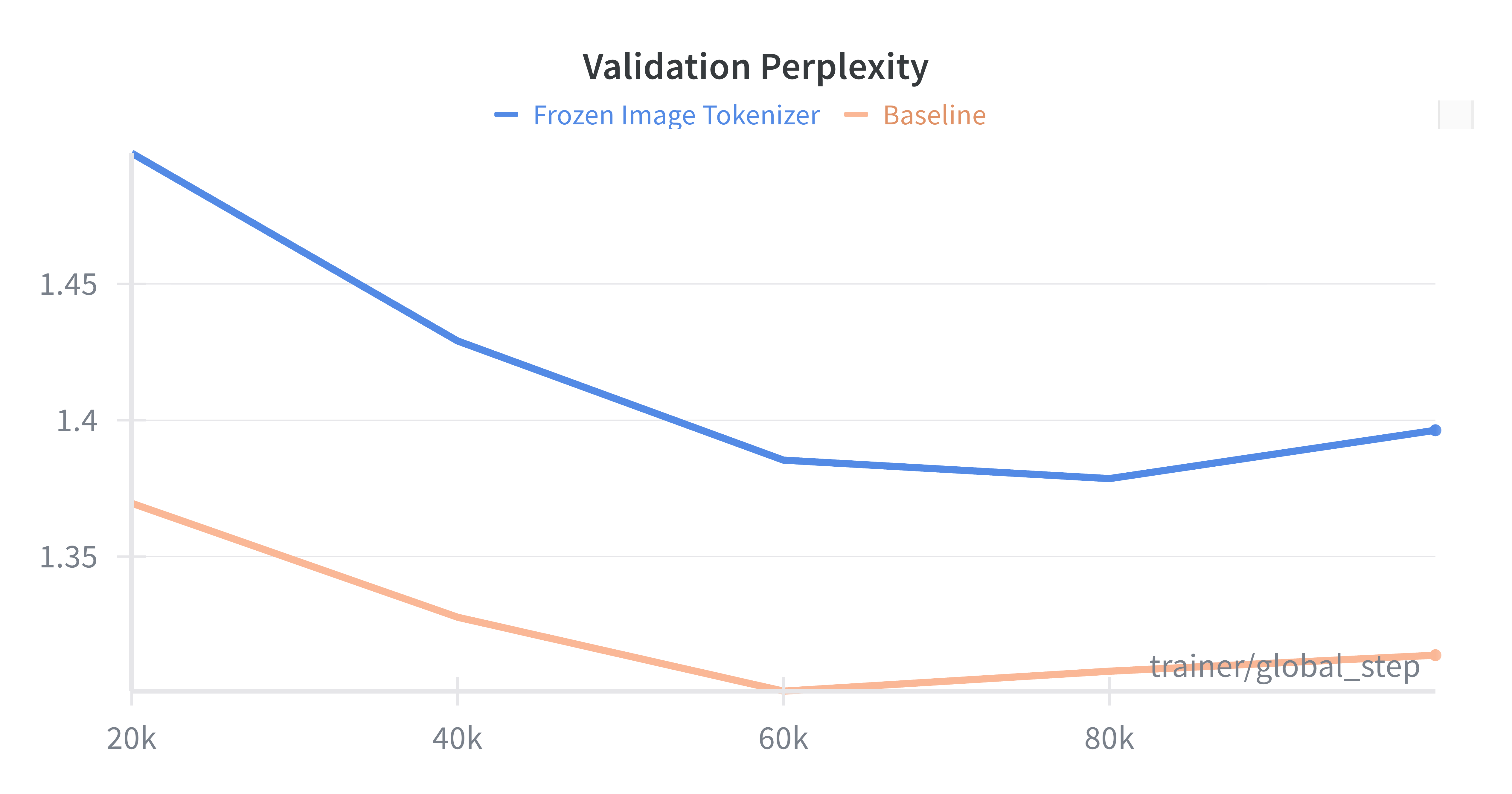}
        \caption{}
        \label{subfig:frozen_valid}
    \end{subfigure}
    \caption{Freezing the pre-trained image tokenizer results in significantly worse training \subref{subfig:frozen_train} and validation \subref{subfig:frozen_train} perplexity. One reason could be that the tokenizers are not pretrained on gaming images so allow them to adapt improves the model.}
    \label{fig:frozen}
\end{figure}

\subsection{Pre-training metrics}

\begin{figure}[H]
    \centering
    \begin{subfigure}{0.48\textwidth}
        \includegraphics[width=\linewidth]{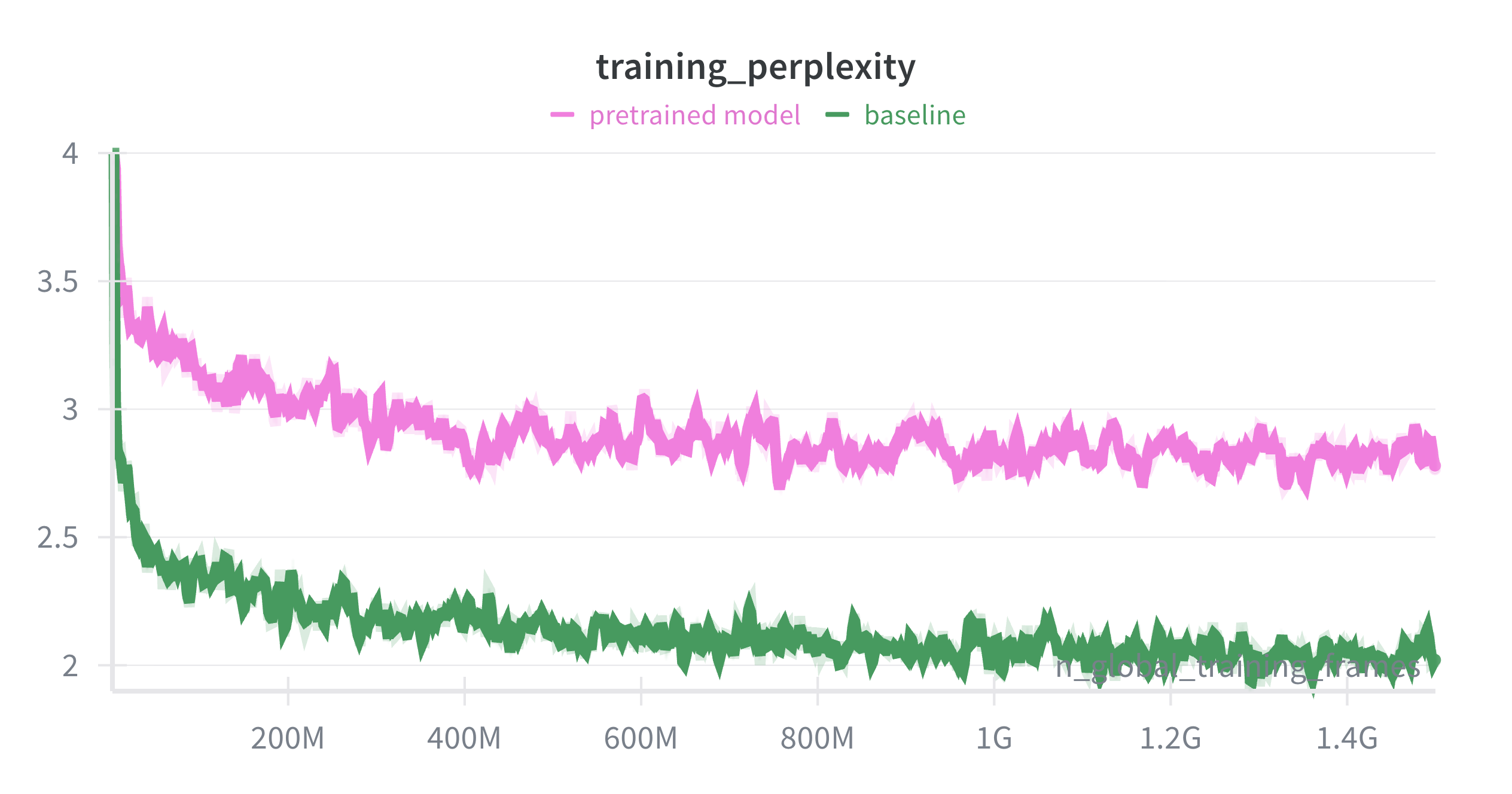}
        \caption{}
        \label{subfig:pretrain}
    \end{subfigure}
    \begin{subfigure}{0.48\textwidth}
        \includegraphics[width=\linewidth]{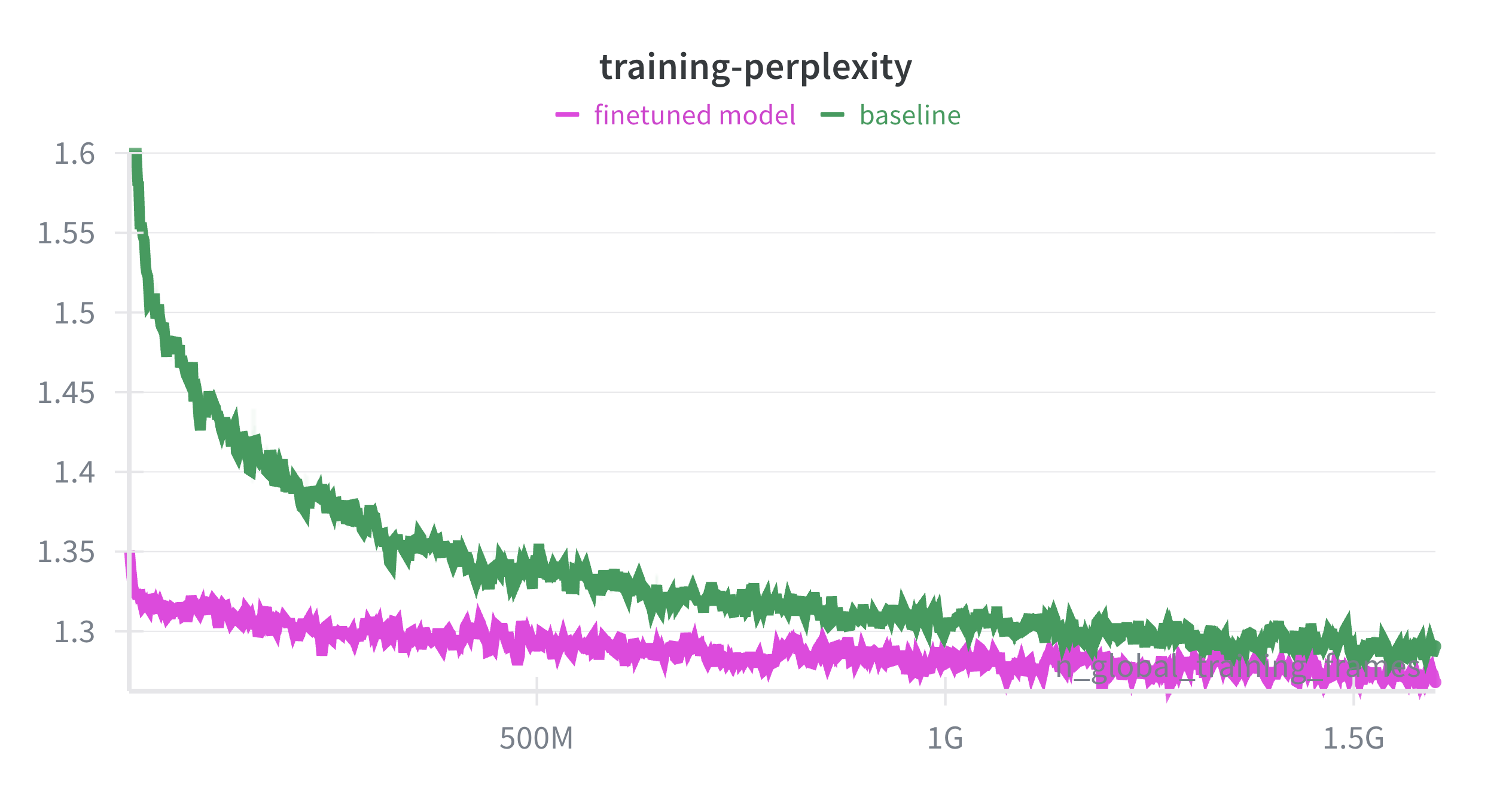}
        \caption{}
        \label{subfig:fine-tune}
    \end{subfigure}
    \caption{Pre-training on unlabelled data using imputed labels results in worse training \subref{subfig:pretrain} and validation perplexity (figure \ref{fig:pretrain-valid}). However, fine-tuning the pretrained model on labelled data improved training perplexity slightly \subref{subfig:fine-tune} and validation perplexity significantly (figure \ref{fig:pretrain-finetune}).}
    \label{fig:pretrain-train}
\end{figure}

\begin{figure}[H] 
    \centering 
    \begin{subfigure}{\textwidth}
        \centering
        \includegraphics[width=0.48\linewidth]{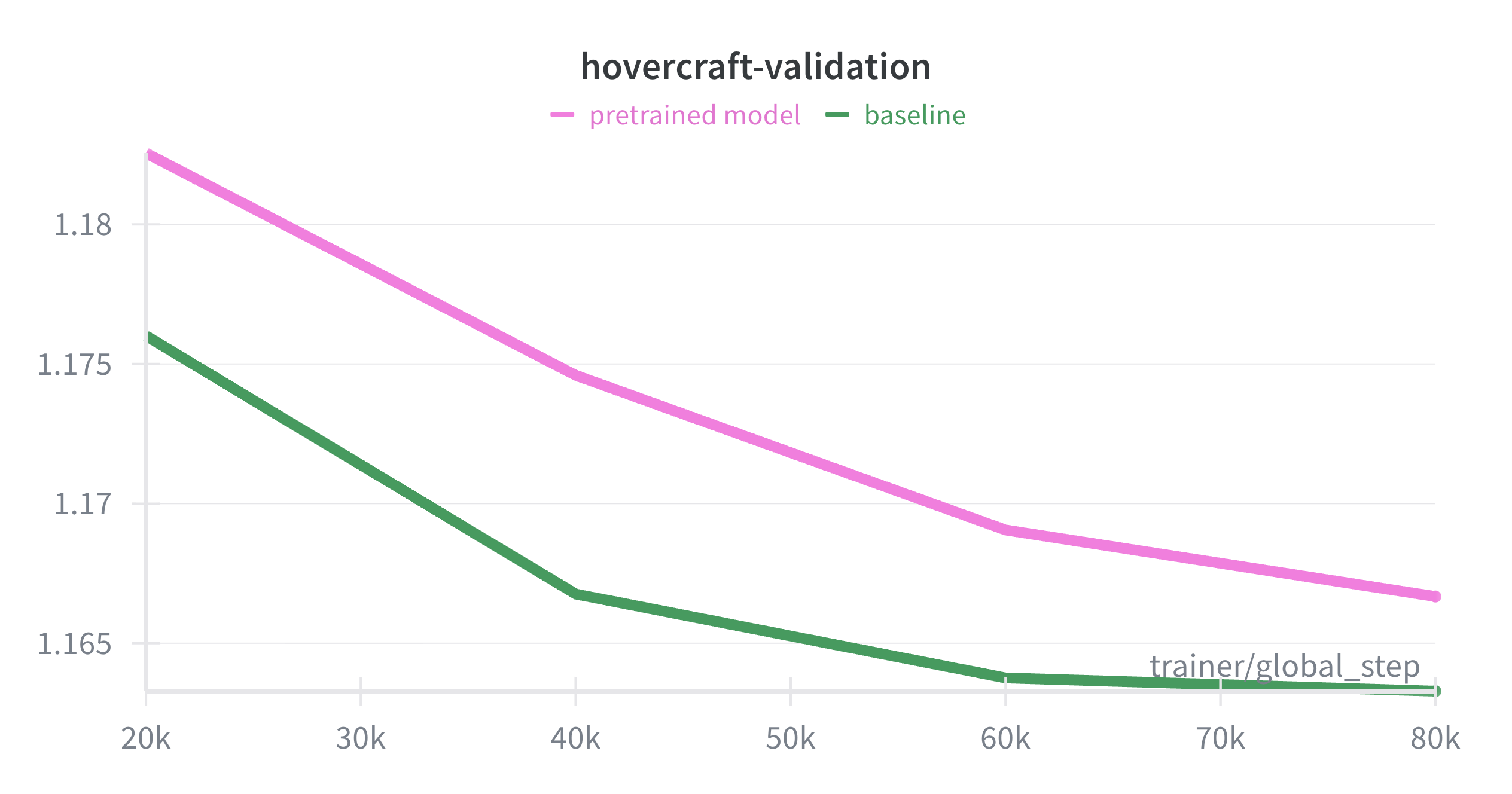}
    \end{subfigure}
    \begin{subfigure}{0.5\textwidth}
        \centering
        \includegraphics[width=\linewidth]{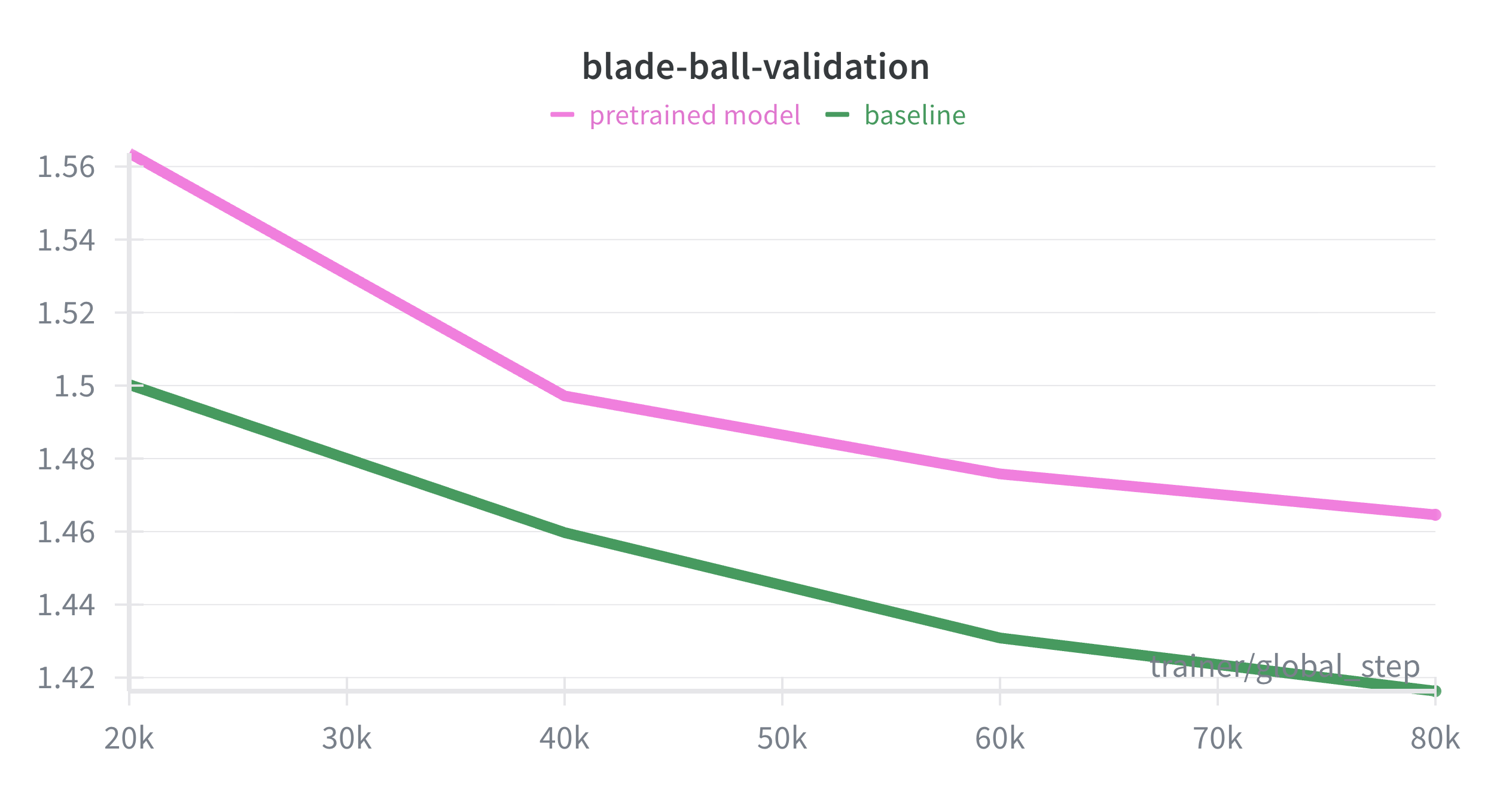}
    \end{subfigure}
    \begin{subfigure}{0.5\textwidth}
        \centering
        \includegraphics[width=\linewidth]{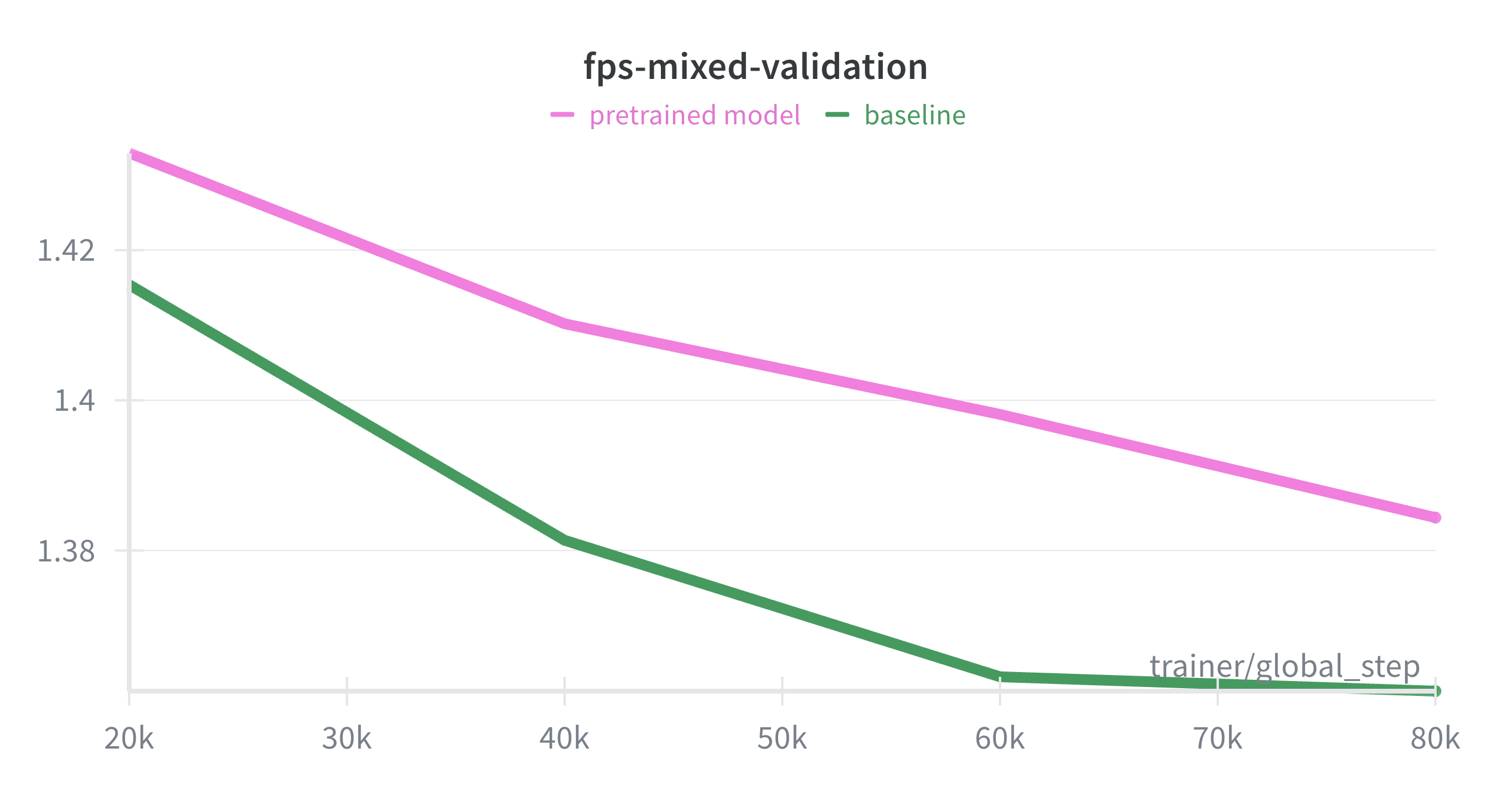}
    \end{subfigure}
    \caption{Validation metrics between pretrained models and a baseline model which is trained only on labeled data on 3 different environments (internal racing game hovercraft, Roblox Blade Ball, a mix of popular first person shooters). We consistently observe across environments the pretrained models (without any fine-tuning stage) with a higher validation perplexity (this is consistent with training perplexity, see figure \ref{fig:pretrain-train}) compared to training only on labelled data.}
    \label{fig:pretrain-valid}
\end{figure}

\subsection{IDM model learns a lower perplexity than a causal policy}

\begin{figure}[H]
    \centering
    \begin{subfigure}{0.45\linewidth}
        \centering
        \includegraphics[width=\linewidth]{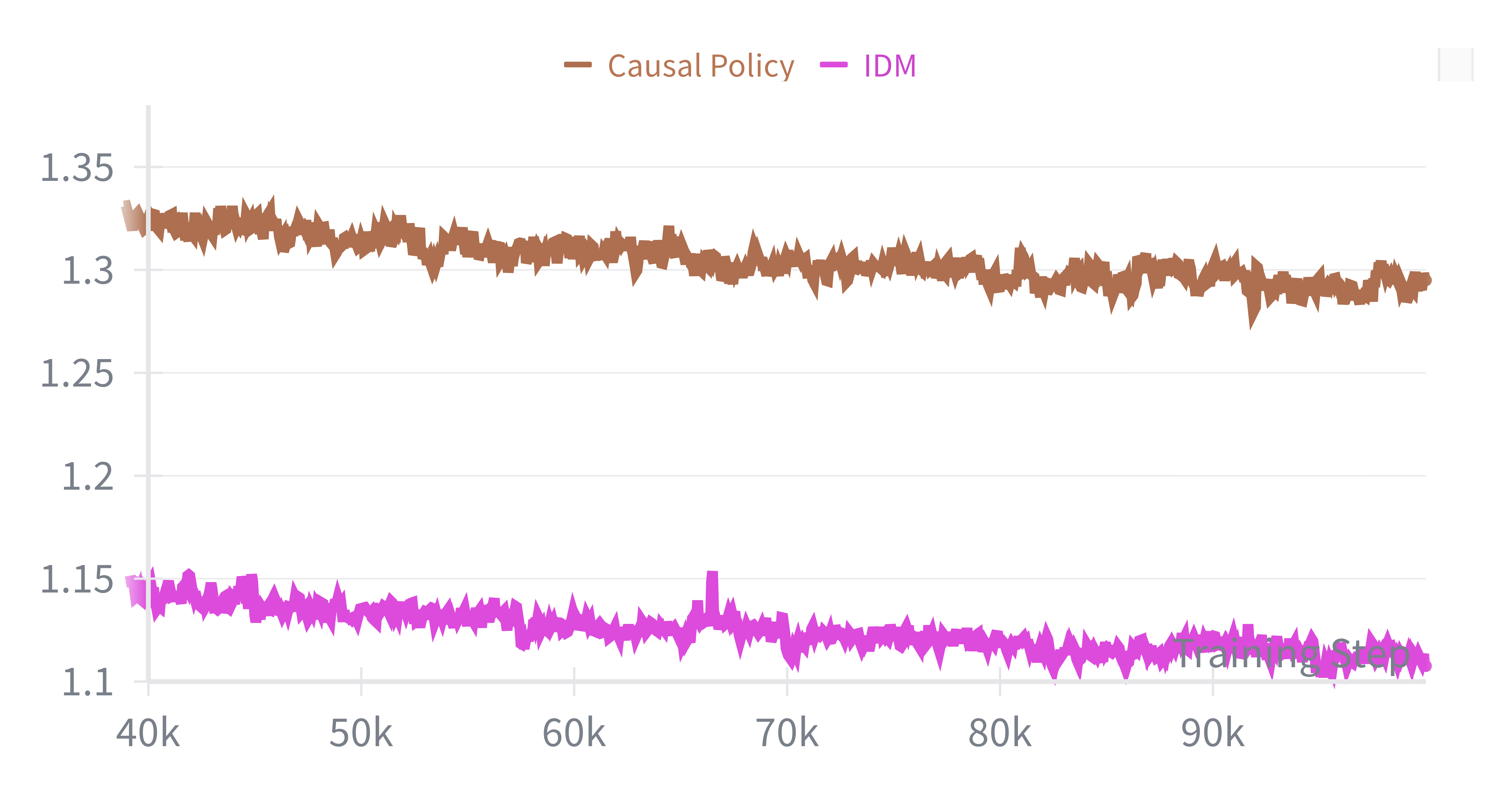}
        \caption{Training}
        \label{fig:idm-train}
    \end{subfigure}
    \hfill
    \begin{subfigure}{0.45\linewidth}
        \centering
        \includegraphics[width=\linewidth]{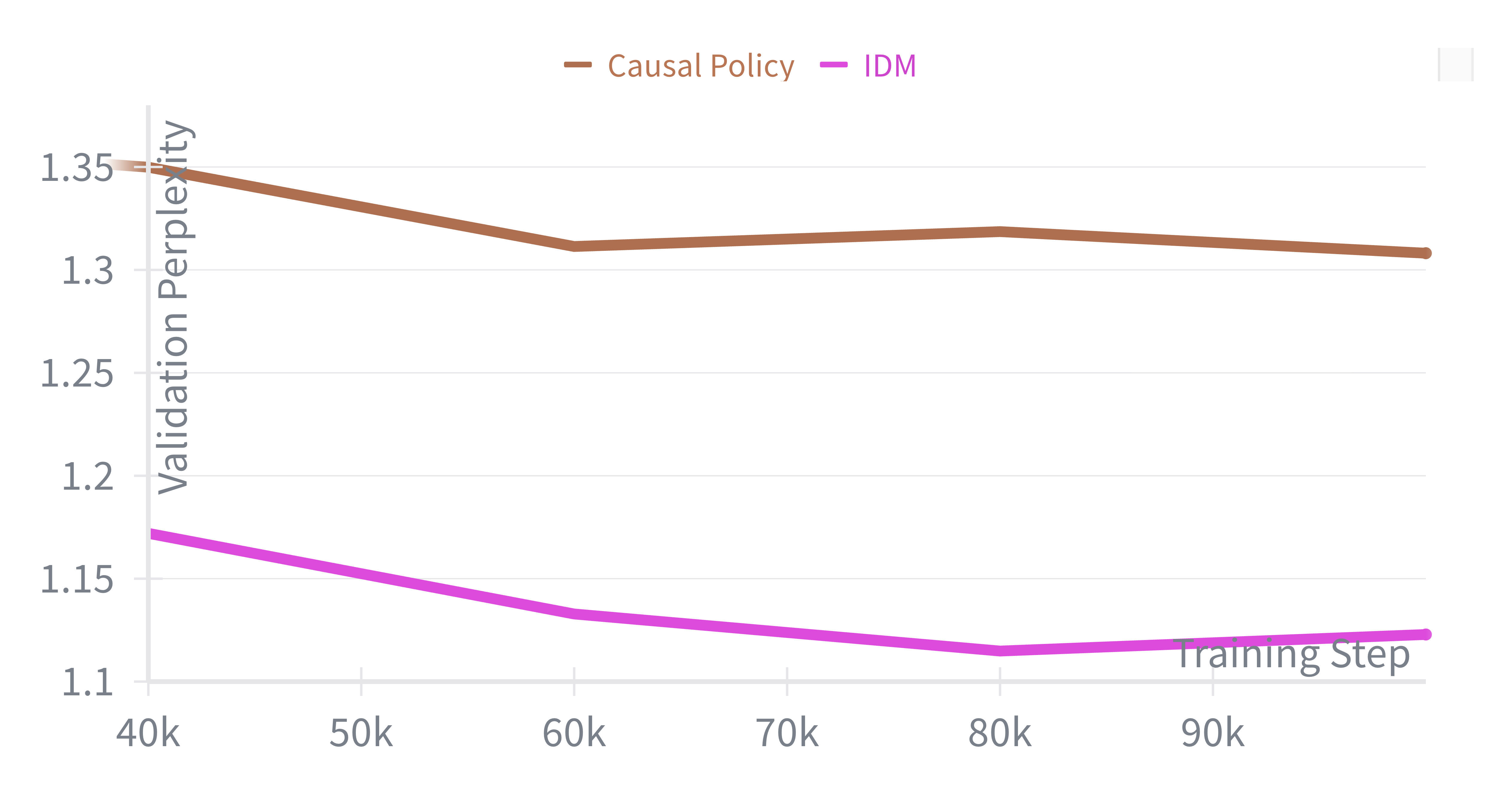}
        \caption{Validation}
        \label{fig:idm-validation}
    \end{subfigure}
    \caption{Training and validation perplexity of a causal policy compared with the IDM. The IDM has a lower perplexity as expected: unlike the causal policy, when predicting which action took place the IDM can make use of information in future frames.}
    \label{fig:idm}
\end{figure}

\subsection{Allowing the policy to observe prior actions harms causality}
\label{appendix:causality}

\begin{figure}[H] 
    \centering 
    \begin{subfigure}{0.48\textwidth}
        \centering
        \includegraphics[width=\linewidth]{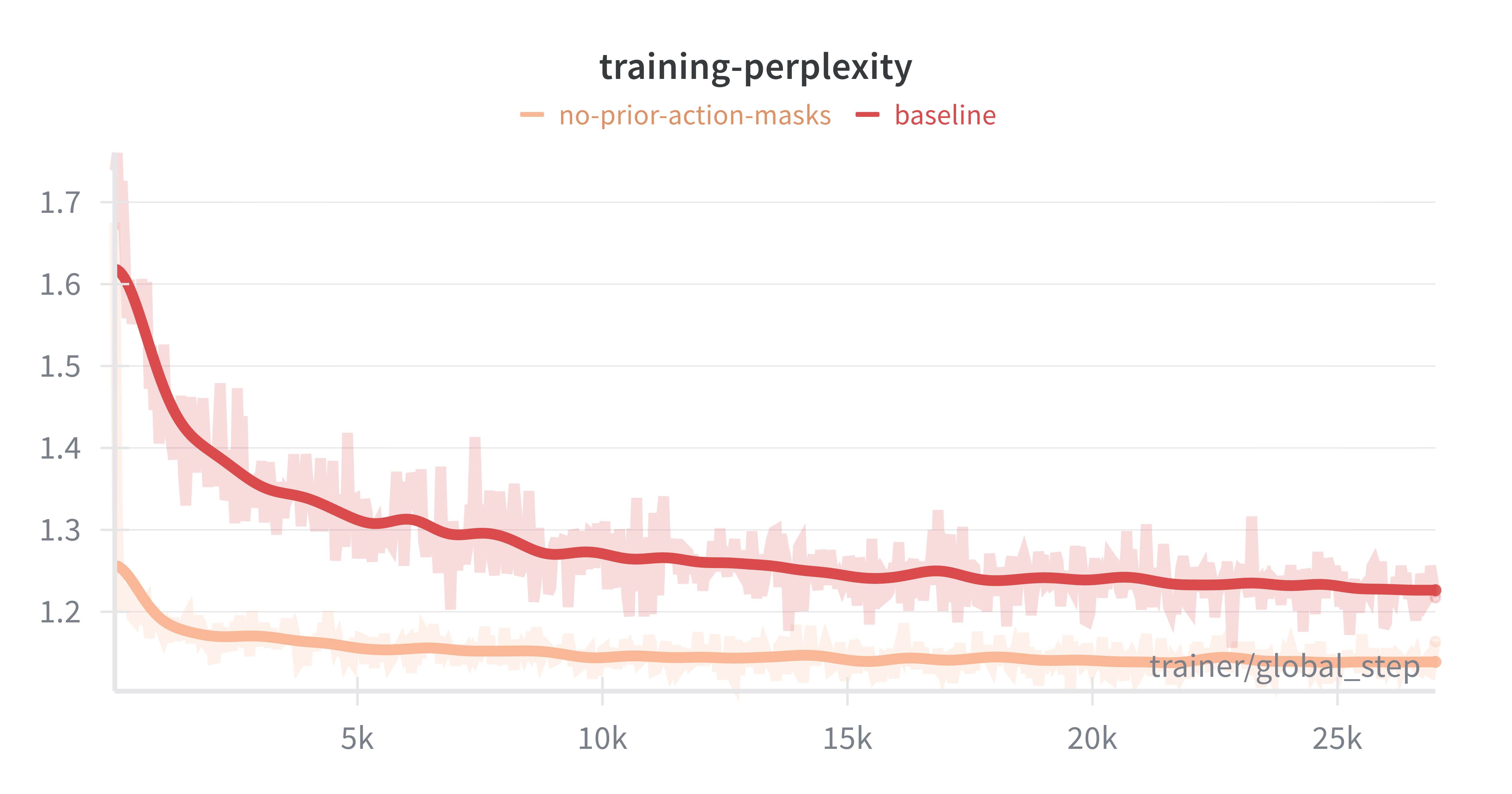}
        \caption{}
        \label{subfig:no-prior-training}
    \end{subfigure}
    \hfill 
    \begin{subfigure}{0.48\textwidth}
        \centering
        \includegraphics[width=\linewidth]{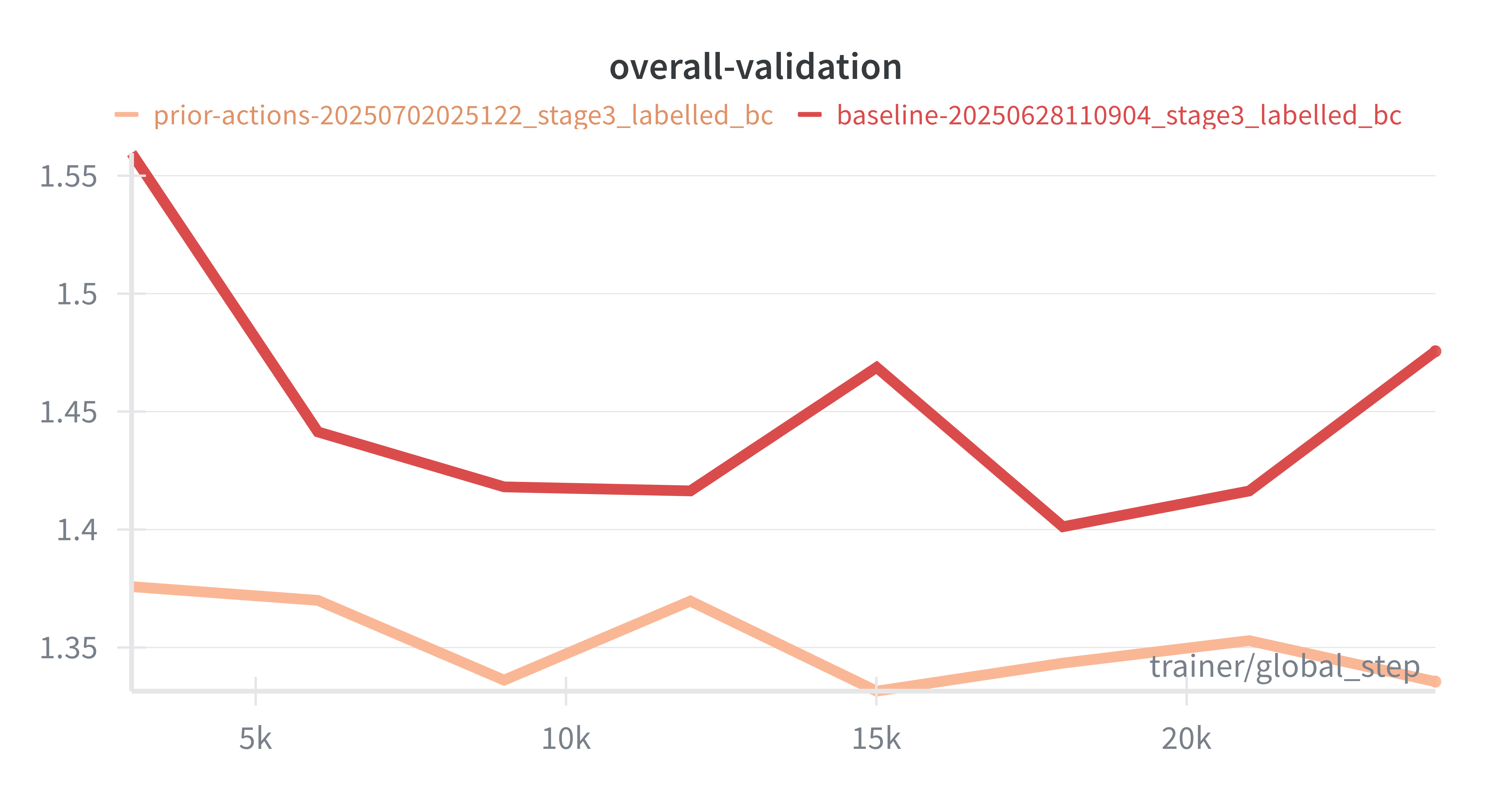}
        \caption{}
        \label{subfig:no-prior-valid}
    \end{subfigure}
    \begin{subfigure}{0.48\textwidth}
        \centering
        \includegraphics[width=\linewidth]{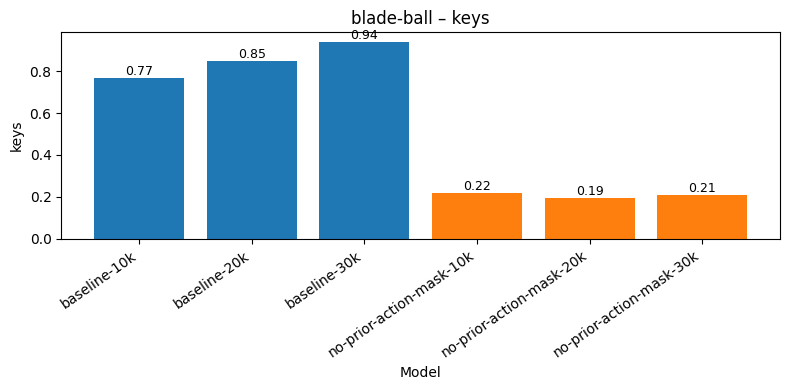}
        \caption{}
        \label{subfig:blade-ball}
    \end{subfigure}
    \hfill 
    \begin{subfigure}{0.48\textwidth}
        \centering
        \includegraphics[width=\linewidth]{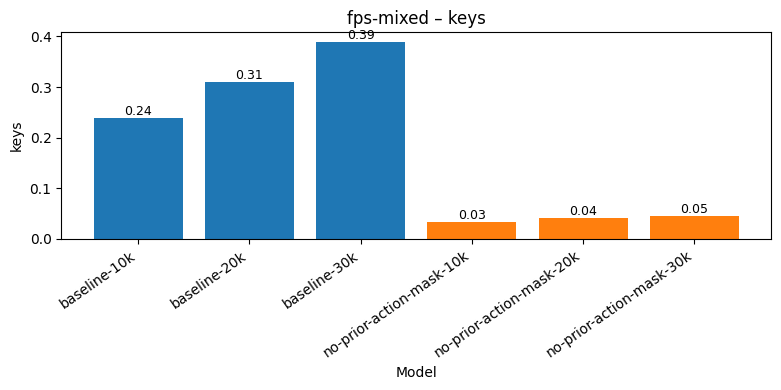}
        \caption{}
        \label{subfig:fps}
    \end{subfigure}
    \caption{When prior actions are unmasked (no-prior-action-masks) both training \subref{subfig:no-prior-training} and validation losses \subref{subfig:no-prior-valid} are lower. However, the model performs extremely poorly in practice. Causal analysis on two games blade-ball (\subref{subfig:blade-ball}) and a variety of first person shooters (\subref{subfig:fps}) provides an explanation. When not masking out prior actions the models tend to learn a non-causal solution by copying the prior action, thus there is little change is action prediction when perturbing the observations. This is resolved by masking out prior actions.
    }
    \label{fig:prior-actions}
\end{figure}

Figure \ref{fig:prior-actions} demonstrates the importance of prior action masking in learning a causal model. Without action masking, the model learns a non-causal solution which performs well on offline metrics (including a validation) but is useless in practice as it largely ignores the content of the observations.

\subsection{Sink token}
\label{appendix:sinktoken}

\begin{figure}[H]
    \centering
    \begin{subfigure}{0.45\textwidth}
        \includegraphics[width=\linewidth]{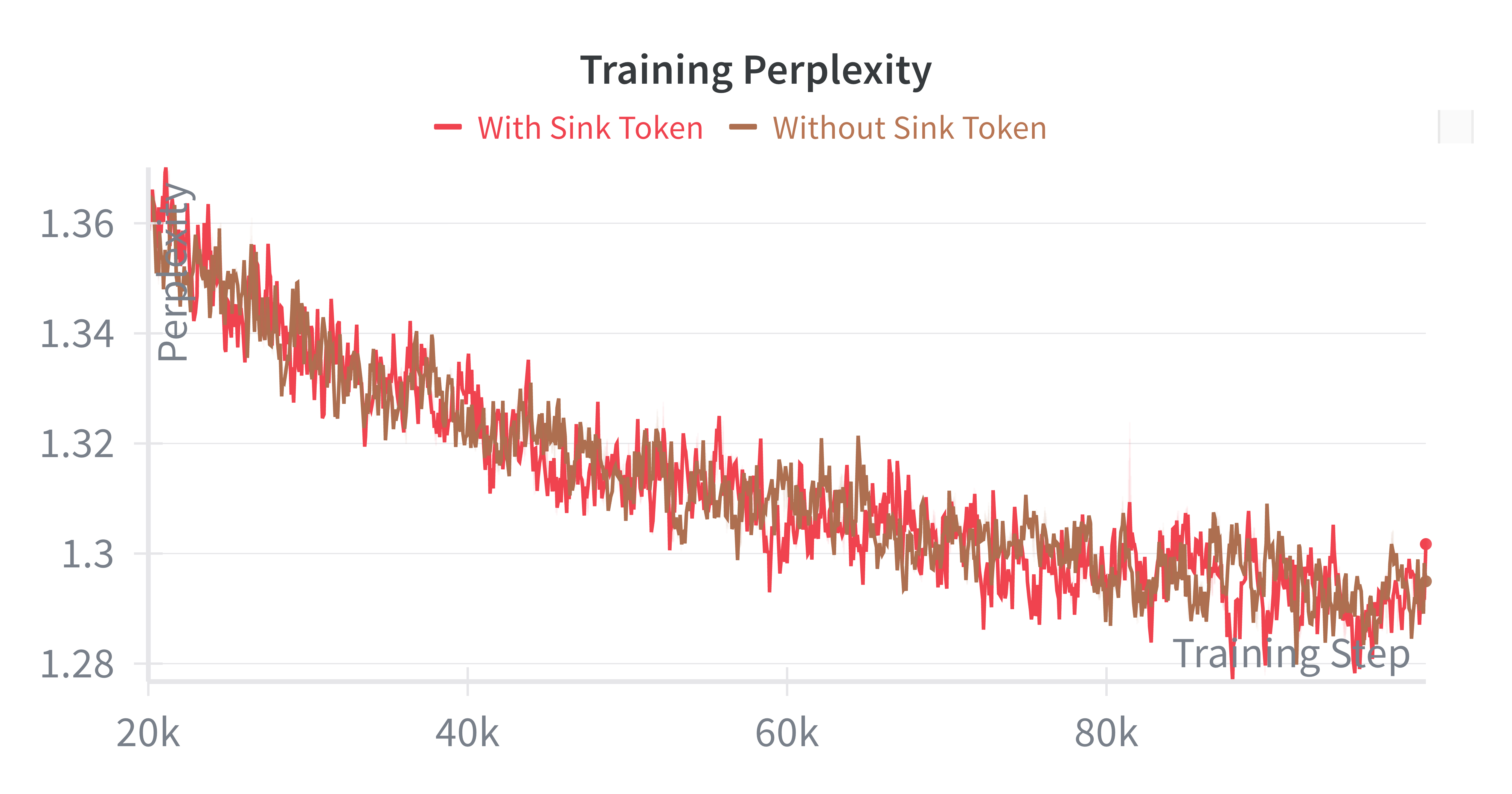}               
        \caption{}
        \label{subfig:sink-train}
    \end{subfigure}
    \begin{subfigure}{0.45\textwidth}
        \includegraphics[width=\linewidth]{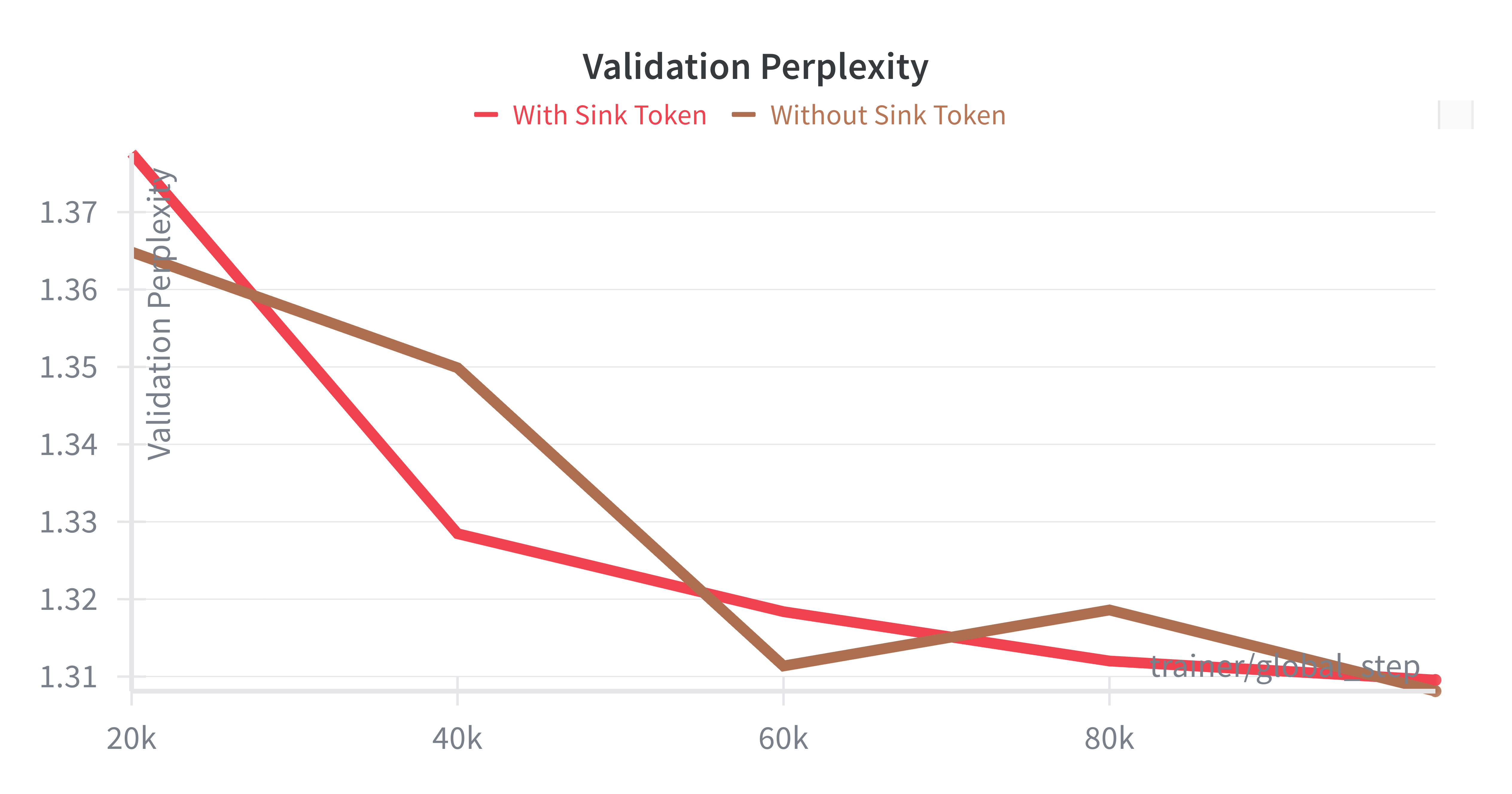}
        \caption{}
        \label{subfig:sink-valid}
    \end{subfigure}
    \caption{Training \subref{subfig:sink-train} and validation \subref{subfig:sink-valid} perplexity comparing the use of a sink token. We find that the use of sink tokens did not appear to improve model metrics.}
    \label{fig:sinktoken}
\end{figure}

We found that using a sink token (see Methods) did not significantly impact performance (figure \ref{fig:sinktoken}).

\subsection{Reasoning Token}
\label{appendix:reasoning}

\begin{figure}[H]
    \centering
    \begin{subfigure}{0.45\textwidth}
        \includegraphics[width=\linewidth]{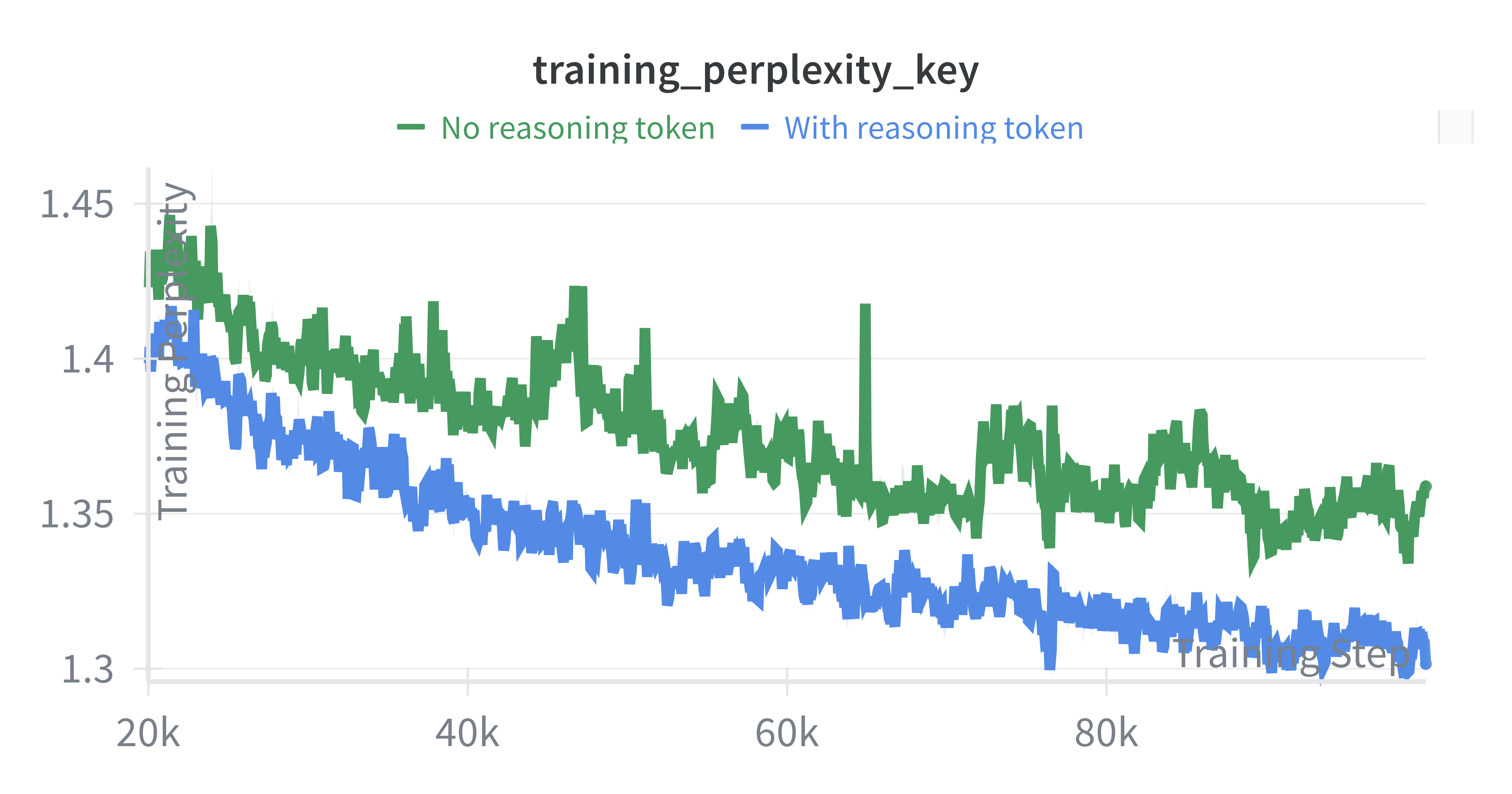}               
        \caption{}
        \label{subfig:think-train}
    \end{subfigure}
    \begin{subfigure}{0.45\textwidth}
        \includegraphics[width=\linewidth]{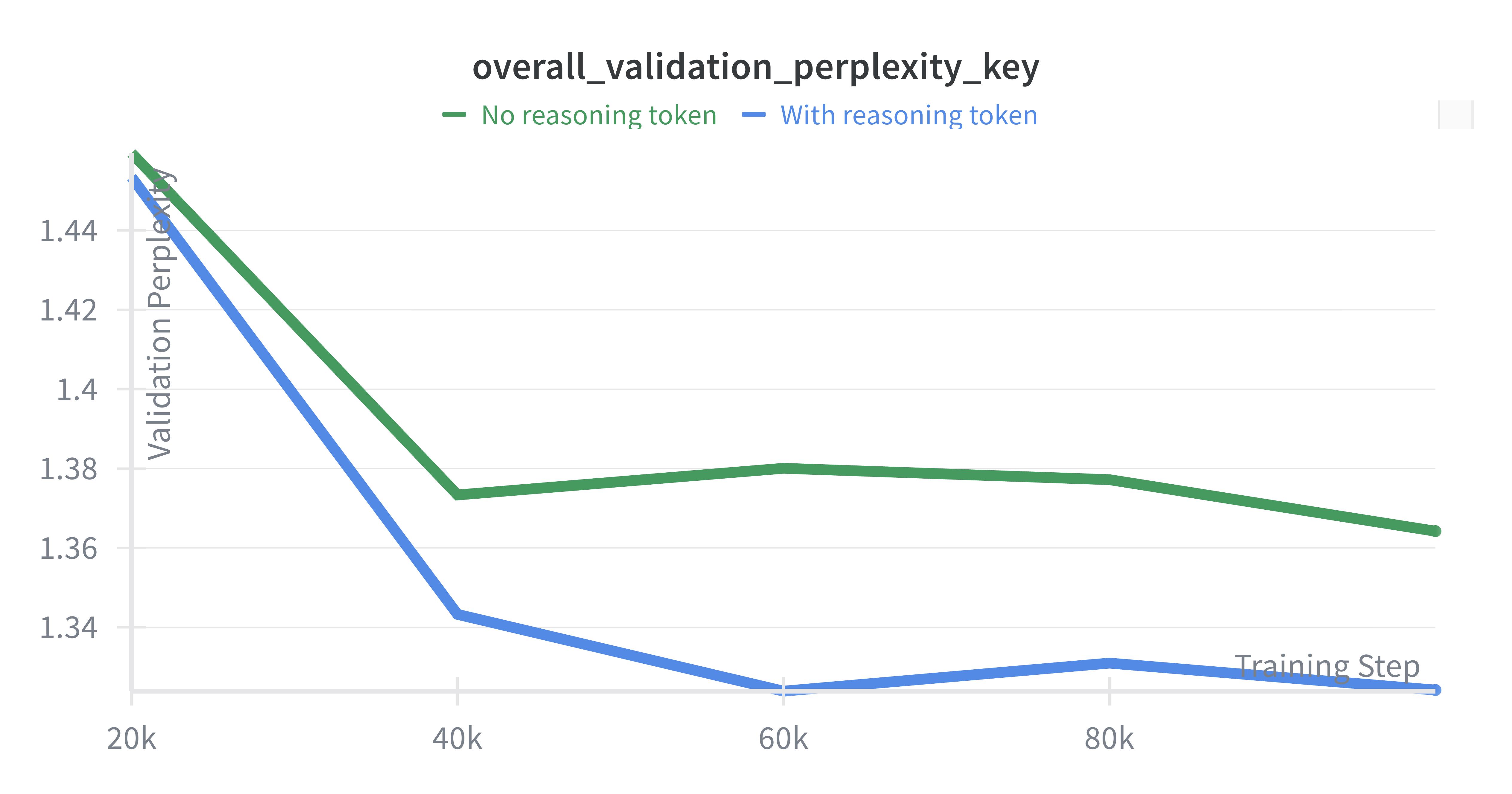}
        \caption{}
        \label{subfig:think-valid}
    \end{subfigure}
    \caption{Training \subref{subfig:think-train} and validation \subref{subfig:think-valid} perplexity comparing the use of a reasoning token. We found both training and validation metrics are improved significantly by allowing the model an additional ``reasoning'' step.}
    \label{fig:thinking}
\end{figure}

We found that the use of a reasoning token between the observation input and action output step (see Methods) significantly improved model performance (figure \ref{fig:thinking}).

\subsection{Image Tokens}

\begin{figure}[H]
    \centering
    \includegraphics[width=0.7\linewidth]{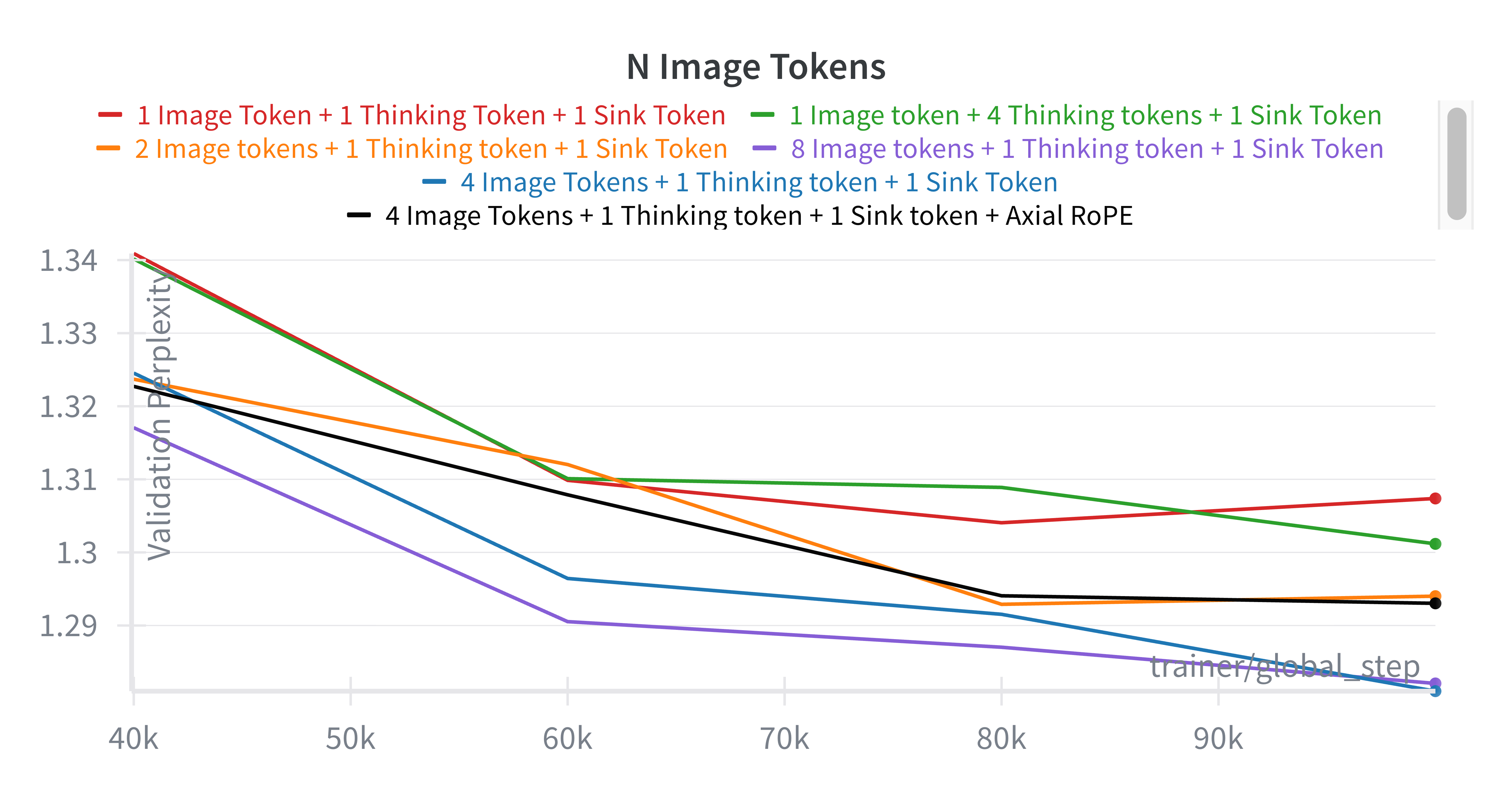}
\caption{Experiments varying the number of image tokens used to represent each image. We find more image tokens improves performance, more so than increasing reasoning tokens.}
\label{fig:n_img_tokens}
\end{figure}

\end{document}